\documentclass{article}

\usepackage{arxiv}

\usepackage[utf8]{inputenc} % allow utf-8 input
\usepackage[T1]{fontenc}    % use 8-bit T1 fonts
\usepackage{hyperref}       % hyperlinks
\usepackage{url}            % simple URL typesetting
\usepackage{booktabs}       % professional-quality tables
\usepackage{amsfonts}       % blackboard math symbols
\usepackage{nicefrac}       % compact symbols for 1/2, etc.
\usepackage{microtype}      % microtypography
\usepackage{lipsum}         % Can be removed after putting your text content
\usepackage{graphicx}
\usepackage{doi}
\usepackage{mathtools}
\usepackage[table,xcdraw]{xcolor}
\usepackage{multirow}
\usepackage{amsmath}
\usepackage{cleveref}       % smart cross-referencing
\usepackage{amsfonts}
\usepackage{bbm}
\usepackage{algorithm}
\usepackage{algorithmic}
\usepackage{amssymb}
\usepackage{adjustbox}
\usepackage{subfig}
\usepackage{polski}
\title{A Survey of Demonstration Learning}

\date{}

\author{ \hspace{1mm}André Correia \\
	NOVA LINCS\\
	Universidade da Beira Interior\\
	Covilhã, Portugal \\
	\texttt{andre.correia@ubi.pt} \\
	\And
	\hspace{1mm}Luís A. Alexandre \\
	NOVA LINCS\\
	Universidade da Beira Interior\\
	Covilhã, Portugal \\
	\texttt{luis.alexandre@ubi.pt} \\
}

\renewcommand{\shorttitle}{A Survey of Demonstration Learning}

\hypersetup{
pdftitle={A Survey of Demonstration Learning},
pdfsubject={q-bio.NC, q-bio.QM},
pdfauthor={André Correia, Luís A. Alexandre},
pdfkeywords={Survey, Demonstration, Learning, Imitation, Demonstrations},
}

\begin{document}
\maketitle

\begin{abstract}
With the fast improvement of machine learning, reinforcement learning (RL) has been used to automate human tasks in different areas. However, training such agents is difficult and restricted to expert users. Moreover, it is mostly limited to simulation environments due to the high cost and safety concerns of interactions in the real world.
Demonstration Learning is a paradigm in which an agent learns to perform a task by imitating the behavior of an expert shown in demonstrations.
It is a relatively recent area in machine learning, but it is gaining significant traction due to having tremendous potential for learning complex behaviors from demonstrations. 
Learning from demonstration accelerates the learning process by improving sample efficiency, while also reducing the effort of the programmer. 
Due to learning without interacting with the environment, demonstration learning would allow the automation of a wide range of real world applications such as robotics and healthcare.
This paper provides a survey of demonstration learning, where we formally introduce the demonstration problem along with its main challenges and provide a comprehensive overview of the process of learning from demonstrations from the creation of the demonstration data set, to learning methods from demonstrations, and optimization by combining demonstration learning with different machine learning methods.
We also review the existing benchmarks and identify their strengths and limitations. Additionally, we discuss the advantages and disadvantages of the paradigm as well as its main applications. Lastly, we discuss our perspective on open problems and research directions for this rapidly growing field.
\end{abstract}

% keywords can be removed
\keywords{Machine Learning, Demonstration Learning, Imitation Learning}
\section{Introduction}
\label{intro}

The need for intelligent systems that mimic human behavior has increased. Future directions in artificial intelligence focus on replacing humans with machines that replicate the desired behavior more consistently. 
Such examples include self-driving vehicles \cite{codevilla} and surgical robots \cite{surgery}.
This push results in continuous advancements in the area of artificial intelligence, where increasingly more difficult problems are solved each year. 
Traditional machine learning methods focus on learning to perform the task while disregarding how the task is accomplished. Copying human behavior implies replicating a sequence of actions that the human would take in different situations. The human behavior can be recorded in the form of demonstration data sets. These data sets can then be used to train a method that replicates the human behavior, by choosing the correct action given the current state of the agent and the environment. 
The solution to this problem involves learning the correct mapping between the states of the environment and the actions of the agent.

This mapping is known in computer science as a policy, a function that selects an action based on the current state.
Traditional programming approaches specify the action for every possible state. Moreover, a different algorithm is required for each task and environment. Furthermore, such algorithms do not scale to high-dimensional environments or continuous action spaces, because specifying the ideal action for all possible state conditions can be tedious or computationally impractical. Hence, these approaches require expertise in the specific area in addition to programming knowledge, which are expensive. 
Because of this, machine learning techniques offer a solution to learn the policies automatically from data.
The common method to learn a policy is through reinforcement learning.
In this family of methods, the agent learns the policy through trial-and-error interactions with the environment.
After each interaction, the agent receives feedback and adjusts its policy accordingly.
The agent often learns super-human policies \cite{deepq}, reaching the task's goal by performing sets of impossible or unlikely actions for a human to perform.
This behavior can be advantageous if the goal is maximum performance or disadvantageous if the agent's goal is to behave naturally.
The clear disadvantage of reinforcement learning approaches is that they require a large number of interactions. 
Even though deep RL has seen immense progress and has tremendous potential it is mostly limited to video games and simulations.
Because the agent learns from failure and tries random actions, the agent and everything in its proximity are potentially unsafe during the agent's learning process if learning is to take place in a real world environment.
This fact prevents its application to the real world such as robotics, and healthcare. Furthermore, reinforcement learning is extremely data inefficient requiring many interactions to converge due to the need for exploration. Hence, using prior data fastens the process.

Because of these reasons, learning from demonstrations is an appealing alternative to reinforcement learning.
In \cite{dqn}, the authors proposed an off-policy RL algorithm that can learn to play Atari games from image data. Two years later, AlphaGo \cite{go} trained the first computer agent capable of beating a professional Go player.
Here, the agent has access to a data set of interactions with the environment performed by an expert teacher.
The agent learns the policy from the demonstrated state-action pairs present in the data set. 
The goal of demonstration learning is to learn complex function approximators like RL, from recorded data, like in supervised learning.
Providing examples of the actions that result in success avoids the need to try incorrect actions during an exploration phase. 
Hence, the agent remains safe during the learning process because it does not have to interact with the environment.
Additionally, the demonstrations ensure that the agent learns the desired behavior exhibited in the data set, avoiding falling into a local minimum, and converges faster than reinforcement learning.
In \cite{deepq}, an agent is pre-trained on demonstration data such that the convergence and its online interactions are safer. Later, \cite{bcq} proposed an algorithm to learn solely from demonstrations, causing the field to gain traction. Alternatively, the demonstrations can be used to learn a dynamics model, which allows the agent to collect new transitions and learn through online RL without having to interact with the environment. Additionally, through Inverse Reinforcement Learning (IRL) the demonstrations can be used to learn a reward function and avoid designing it, which is difficult in high-dimensional domains. 

However, the agent is completely reliant on the demonstrations for learning. The demonstrations must cover the state space for the agent to learn the task. Depending on the task, this can be difficult and expensive due to the size of the state and action spaces. Demonstration learning methods attempt to generalize to unseen states. However, unlike supervised learning, the data is not i.i.d, which makes generalization difficult. Then, if the agent encounters states outside the distribution of the data set it likely fails and such a failure has serious consequences in the real world. Because of this, the demonstration learning methods try to mitigate the distributional shift, which is still an open problem.
Furthermore, recording demonstrations performed by a human requires finding a way to record the state-action pairs in the demonstrations, showing good behavior, and covering various settings. The demonstrations can suffer from noise in the sensors used to capture the demonstrations, inaccuracies, or inconsistencies of the demonstrator. 
Therefore, demonstration learning approaches avoid the direct copy of the demonstrated behavior and attempt to generalize to non-demonstrated trajectories. The usage of demonstration data sets is the primary source of differentiation between demonstration learning methods.
The usage of demonstrations reduces the programming overhead, opening the doors for non-experts. Consequently, this area has seen an exponential increase in interest due to its tremendous potential.
After learning the policy from demonstrations, it can be refined with online interactions through RL, with the benefit that the initial policy is safer than an initial random RL policy.

\begin{figure}[!t]
\centering
  \includegraphics[width=0.7\textwidth]{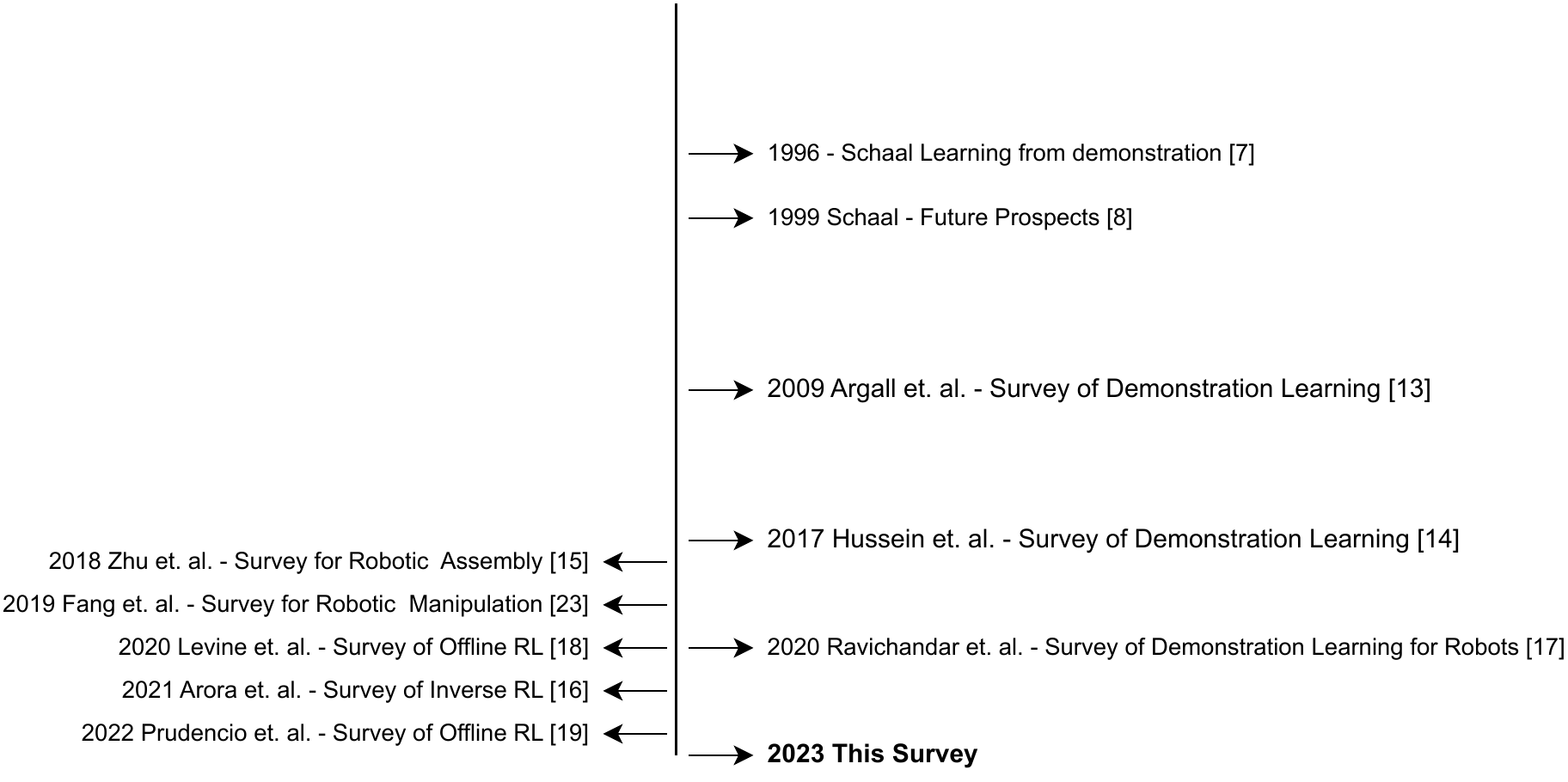}
    \caption{Timeline of surveys. General demonstration learning surveys are presented on the right side, while surveys specific to a sub-area or application are presented on the left side.}
\label{surveyfig}
\end{figure}

The paradigm was created in the eighties and has since been viewed as the future of robotics \cite{schaal1996learning,SchaalFuture}.
Since then, it has been applied to teach a plethora of tasks and to a variety of robots. Such applications include aerial and grounded navigation \cite{NASA}, video games \cite{deepq}, and controlling different kinds of robots ranging from manipulators \cite{knot} to humanoids \cite{speert}. 
Due to deep learning, research interest in teaching robots through demonstrations has grown exponentially over the past two decades, resulting in significant growth in the number of publications.
This fast growth resulted in the creation of various surveys.
The first surveys review the beginning of the history of the paradigm and early attempts to teach robots from demonstrations.
One presented an overview of demonstration learning and defined the problem using four core questions for the field: how, what, when, and whom to imitate  \cite{SchaalBillard}.
In a following survey, \cite{senhoras}, the authors address different design choices and propose a categorization for the field.
Later, \cite{surveyverboso} focuses on reviewing artificial intelligence methods which are used to estimate the policies.
\cite{surveyzhu} reviews recent research and development in the field with the main focus being how to demonstrate behaviors to assembly robots, and how to extract the manipulation features. \cite{surveyirl} provides a survey of inverse reinforcement learning.
In \cite{recentadv}, the authors provide an overview of the collection of machine-learning methods and show their respective advantages and disadvantages. Next, \cite{levinesurvey} and later \cite{surveryofrl}, provide a review of offline reinforcement learning methods. The timeline of published surveys is represented in Fig. \ref{surveyfig}.

Even though the surveys mentioned above aimed at restricting the terminology, it remains diverse in papers published in recent years. Demonstration learning, imitation learning, behavioral cloning, and offline reinforcement learning are popular terms used to describe the same paradigm. For consistency, this survey uses demonstration learning to refer to the paradigm.

Due to the rapid growth of the field, there is a need for a new survey. This survey provides an overview of the steps required to learn from demonstrations and the different methods employed by researchers in each step. The reviewed literature covers a variety of applications. The survey explains each step in a general way such that it can be applied to most tasks. The following section \ref{problem} presents a formal definition of the demonstration learning problem. Section \ref{data set}, discusses the methods for collecting demonstration data and creating the data sets. Next, in section \ref{policy}, the learning methods available from the demonstration data are explained. Then, we present the available benchmarks in section \ref{benchmarks}. Next, in section \ref{applications} the main applications of demonstration learning are listed, followed by the advantages and disadvantages of the paradigm in section \ref{advdis}. Lastly, in section \ref{future} the future directions that demonstration learning research should tackle are discussed, ending with a conclusion in section \ref{conclusion}.

\section{Problem Definition}
\label{problem}

\begin{figure}[!t]
\centering
  \includegraphics[width=0.5\textwidth]{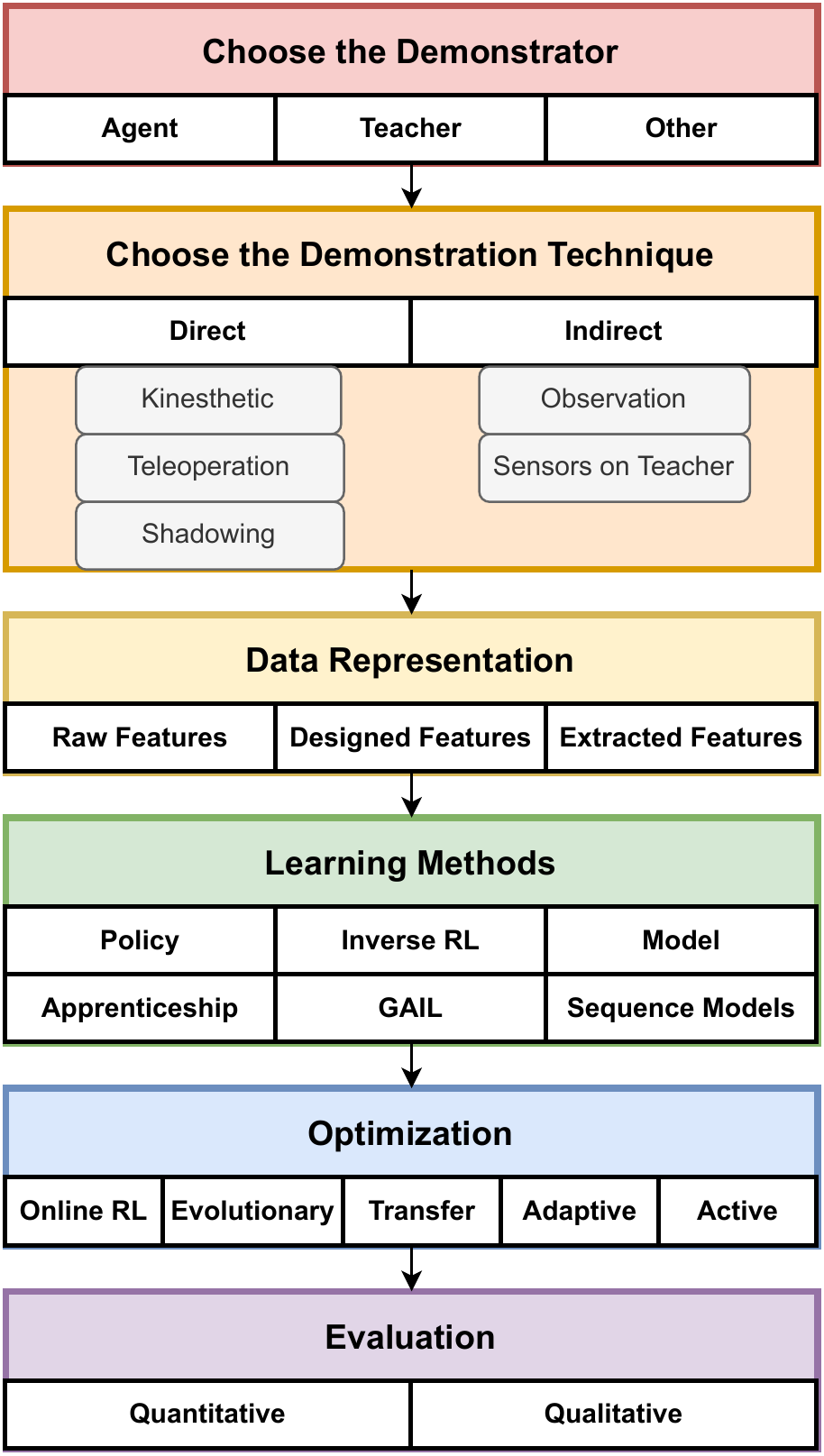}
    \caption{Demonstration learning flowchart.}
\label{diagramfig}
\end{figure}

This section explains the demonstration learning problem and several relevant concepts. The overall sequence of steps in demonstration learning is represented in figure \ref{diagramfig}. 

An agent is an entity that autonomously interacts within an
environment toward achieving or optimizing a goal \cite{russel}. It receives information from the environment through its sensors and interacts with the environment using its actuators, based on its policy.

Demonstration learning is a mixture of supervised learning and reinforcement learning. In supervised learning, the agent receives the labeled training data and learns an approximation to the function that produced the data. In reinforcement learning, the agent must collect interaction data to learn from, by interacting with the environment through trial and error. In demonstration learning, the training data is a set of environment interactions collected beforehand by a teacher executing a task. 
The goal of demonstration learning is to have an agent perform a task by learning from interactions demonstrated by an expert demonstrator and recorded in a data set.

Demonstration learning expands from the reinforcement learning paradigm commonly defined as a Markov Decision Process (MDP), formulated by the tuple $<S, A, P, \Delta_0, R, \lambda>$ \cite{barto}. 
An MDP is a mathematical formulation that enables the creation of theoretical statements and proofs in RL, where $S$ is the set of the possible environment states, $s \in S$, and $\Delta_0$ denotes the initial state distribution. At each state, the agent can choose an action $a \in A$ from the set of possible actions. Acting changes the state of the environment. The mapping between states through the actions is defined by the state transition function $P(s_{t+1}\mid s_t, a_t) : S \times A \rightarrow S$. 
The Markov property determines that the transition function completely defines the dynamics of the environment. That is, the probability of state $s_{t+1}$, only depends on the current state $s_t$ and selected action $a_t$, regardless of past transitions. 
The policy $\pi : S \rightarrow A$ is a function that selects an action given a state of the environment. A more common definition is to formulate the policy as a probability distribution $\pi(a_t\mid s_t)$, where the policy returns the probability of taking action $a_t$ given the agent is at the current state $s_t$.
The correct selection of the action for any given state is what allows the agent to perform the task. After interacting with the environment, the agent receives a reward $r_t = R(s_t, a_t, s_{t+1})$, which indicates the quality of the interaction. In reinforcement learning, the policy is optimized to maximize the expected future rewards $\mathbb{E}[\sum_{t=0}^{\infty} \lambda R(s_t, a_t, s_{t+1})]$.
A trajectory is a sequence of $H+1$ states and $H$ actions and rewards $\tau = (s_0, a_0, r_0, ..., s_H)$, where $H$ is the episode's horizon, which may be infinite in the case of non-episodic environments.
With these definitions, the probability density function for a given trajectory $\tau$ under the policy $\pi$ is $\rho_\pi(\tau) = \Delta_0 (s_0) \prod_{t=0}^{H-1} \pi(a_t\mid s_t)P(s_{t+1}\mid s_t, a_t)$. Lastly, $\lambda$ is the discount factor.

In some settings, we do not have access to the full state information of the environment and have to work with observations $o_t \in O$. This formulation is named partially-observable MDP (POMDP), defined by the tuple $<S, A, O, P, \Delta_0, E, R, \lambda>$, where $O$ is the set of observations and $E(o_t \mid s_t)$ is the function that maps states to observations. To counter the limitations of learning from observations, methods combine consecutive past observations to provide the policy with time-varying information such as velocity and direction.

One of the main reasons behind the success of machine learning methods is the usage of large data sets. However, in RL the data set is collected during training and can be expensive and unsafe to collect in the real-world.

In demonstration learning, the agent has access to a data set of $N$ demonstrations $D_{demo} = \{ \tau_i, i \in [0,N[ \}$. Each demonstration is the sequence of visited states and the respective actions chosen by the expert demonstrator $\tau_i = {(s_t, a_t, s_{t+1}, t \in [0,L[)}$, where $L$ is the length of the sequence. The policy is estimated from the behaviors shown in the demonstration data set. The agent learns by increasingly better imitating the behavior of the teacher represented in the demonstrations. Therefore, the behavior of the agent should converge to a working and intended behavior.

Behavior cloning is the family of methods where the policy is trained to output the demonstrated action for a given state. Hence, the problem becomes a classification problem for discrete action spaces or a regression problem for continuous action spaces. However, the quality of the learned behavior is limited to the ones present in the demonstrations. Because of this, demonstrations can be used to formulate a reward function. This family of methods is called inverse reinforcement learning. Here, the agent is rewarded for how similar the action is to the one in the data set for a given state. Alternatively, the demonstrations can include the environment rewards in addition to the states and actions:
$\tau_i = {(s_t, a_t, r_t, s_{t+1}, t \in [0,L[)}$. This family of methods is named offline reinforcement learning because the agent has access to the interaction data in an offline manner.
Here the ideal policy $\pi*$ is estimated by maximizing the expected accumulated reward for all trajectories $\tau$:
$\pi* = \arg\max_\pi \mathbb{E}_{\tau \sim \rho_\pi(.)}[R_\tau]$, where $R_{\tau} = \sum_{t=0}^{N-1} \gamma^{t} r_t$ is the discounted accumulated reward of trajectory $\tau$ with $N$ transitions.
Most methods estimate a state-value function to optimize the policy: $V^\pi(s_t) = \mathbb{E}_{\tau \sim \rho_{\pi}(.\mid s_t)}[R_\tau]$, which maps a state to the expected return when starting from that state. Similarly, an action-value function $Q^\pi(s_t,a_t)$, maps state-action pairs to the expected return starting from state $s_t$, and using action $a_t$.

Under offline RL, the goal is to use the data sets to generalize instead of naive imitation learning by finding the good parts of the demonstrated behavior. Even if the data set has bad behaviors, finding the good parts would result in an improvement over the demonstrations. Though distinguishing bad from good behaviors is difficult, offline RL accounts for long-term consequences of immediate actions through the value-function, unlike behavior cloning.

\section{Demonstration Data Set}
\label{data set}
The first step to learning from demonstrations is the creation of the demonstration data set. The data set is a set of demonstrations made by an expert teacher, each corresponding to a sequence of state-action pairs. As stated previously, the data set can include extra information such as the environment's rewards for each interaction.
The developer has a plethora of design options for the system. This section presents the different options for designing the demonstration data set, their impacts on the final design, and their advantages and disadvantages compared to the other choices.

The first step is selecting the demonstrator. The next step is selecting the demonstration technique which depends on the type of demonstrator selected. Then, it is important to define how the data is stored. The data must be usable by the learning agent. Ideally, the pairs in the data set map directly to state-action pairs usable by the learner. However, this is not always possible, and conversion mechanisms may need to be applied. For example, learning from images requires the extraction of features, which may be manually-designed or learned. The demonstrated images are captured from the point of view of the teacher and may not be directly usable by the learning agent. Challenges that arise from the differences between the contexts of the demonstrator and the learner are named correspondence issues by \cite{correspondence}. All these steps are explored in the following sections.

\subsection{Choosing the Demonstrator}

Two choices have to be made regarding the demonstrator: selecting who controls the demonstration and who executes the demonstration. 
A human, or an agent, different from the learning agent, can demonstrate the task. 
In both cases, the demonstration is controlled and executed by someone other than the learning agent.
However, through teleoperation, a human can control the agent and have it demonstrate the task. Here, a human teacher controls the demonstration while the learning agent executes the demonstration.

The choice of the demonstrator is critical for the success of the system. It determines what algorithms can be used. Suppose the learning agent itself performs the demonstrations, for example through teleoperation. In that case, the learner's state-action spaces will automatically match with the spaces in the data set. However, if the agent executing the demonstrations is different from the learning agent, the state and action spaces represented in the demonstrations will likely need to be mapped to the learner's spaces.

\subsection{Demonstrator and Learner Matching}

\begin{figure}[!t]
\centering
  \includegraphics[width=0.8\textwidth]{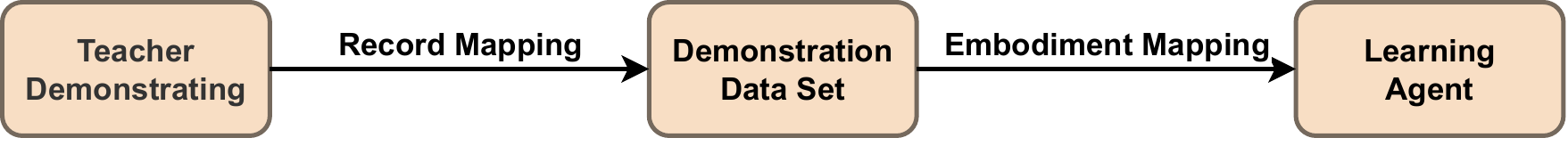}
    \caption{Record and Embodiment Mapping as per \cite{senhoras}.}
\label{mappingsfig}
\end{figure}

This subsection deals with matching the state-action pairs performed by the teacher represented in the demonstration data set, with the learner's state-action pairs, which allow it to perform the task.
In \cite{senhoras}, the authors defined two types of mappings: record mapping and embodiment mapping. This notation is not actively used in the literature. However, we choose to adapt it here as we find it useful to classify the different demonstration techniques. Record mapping corresponds to the mapping between the teacher's demonstrated state-action pairs and the recorded state-action pairs in the data set $(s_{recorded}, a_{recorded}) = m_R(s_{teacher}, a_{teacher})$. Embodiment mapping corresponds to the mapping between the state-action pairs recorded in the data set and the state-action pairs performed by the learning agent $(s_{agent}, a_{agent}) = m_E(s_{recorded}, a_{recorded})$. These mappings are represented in figure \ref{mappingsfig}.
Each mapping corresponds to applying a conversion function to the input to produce the output. Neither mapping changes the information of the demonstration data set. Instead, it only changes data from the context of the teacher to the context of the agent. 

Ideally, the data is recorded by the learning agent and no mapping is required. In such cases, these mappings are equivalent to using the identity function $I(s, a)$. However, in some scenarios, the demonstrator's setting cannot be directly recorded and/or directly applied to the learner. Hence, one or both conversion mappings need to be created. 
For any given problem, the addition of each mapping adds a potential source of errors. Because of this, the more mappings that are applied, the more difficult it is to translate and reproduce the teacher's original behavior.

As an example, consider a human teacher demonstrating a task to a robot using its own body. A camera records the demonstrations in the form of images. An implicit record mapping is applied to the information captured by the camera to generate the images because the images do not contain the entire information of the environment. Additionally, the actions present in the images are unknown to the robot. The robot does not know what action was performed by the teacher that converts the information in a frame to the information in the next frame. Therefore, embodiment mapping needs to be used to generate state-action pairs that the robot can use.

\subsection{Choosing the Demonstration Technique}
In this section, we discuss the different techniques for acquiring demonstrations.
\cite{senhoras} categorize demonstration techniques into two categories, demonstration, and imitation, based on their requirement for the embodiment mapping. 
We adopt a simpler categorization proposed in \cite{surveysmall}, which divides the demonstration techniques into indirect demonstration and direct demonstration. In direct demonstration techniques, the learning agent performs the demonstration, while in indirect, an external agent demonstrates the task. 

In the direct category, there is no embodiment mapping because the demonstration is performed directly on the target agent. Contrarily, in indirect techniques, there is an embodiment mapping because the demonstration is not performed on the learning agent. They then further group the approaches within each of the two categories based on the requirement of record mapping. This categorization is presented in Table \ref{tab:tabdemotec}.

The demonstration techniques involve choosing how the demonstration data is provided to the learner. The most common approach is to have the data available beforehand and learn the policy from a data set. Contrarily, the data can become available during training, resulting in incremental policy updates. Such approaches are usually employed in response to the performance of the policy during training. For example in \cite{dogged}, the agent shows its confidence score for the chosen action for a given state. The teacher can choose to intervene and demonstrate or accept the action.

\cite{Chernova} covers techniques that allow a robot to refine the existing model, mainly through interactive and active learning approaches. Interactive task learning was proposed in \cite{ITL}, where the agent actively tries to learn a task through natural interactions with a human instructor. 
The idea of modeling verbal and non-verbal cues that the teacher uses during the teaching process to improve learning is explored in\cite{Cues}. The idea of asking for help while learning is explored in \cite{askforhelp}. Here a data-driven method was created to estimate the human's beliefs after hearing the request and create better requests that lead people toward useful help. We provide a categorization of the available demonstration techniques in Table \ref{tab:tabdemotec}.

\begin{table}[!t]
%\centering
\caption{Categorization of the demonstration techniques.\label{tab:tabdemotec}}
\resizebox{\textwidth}{!}{
\begin{tabular}{c|ccc|cc}
                                 & \multicolumn{3}{c|}{\textbf{Direct}}                                                                                     & \multicolumn{2}{c}{\textbf{Indirect}}                           \\ \cline{2-6} 
\textbf{}                        & \multicolumn{1}{c|}{Teleoperation}            & \multicolumn{1}{c|}{Kinesthetic}              & Shadowing                & \multicolumn{1}{c|}{Observation}           & Sensors on Teacher \\ \hline
\textbf{Mapping required}        & \multicolumn{1}{c|}{No}                       & \multicolumn{1}{c|}{No}                       & Record                   & \multicolumn{1}{c|}{Embodiment and Record} & Embodiment         \\
\textbf{Demonstration performer} & \multicolumn{1}{c|}{Learner}                  & \multicolumn{1}{c|}{Learner}                  & Learner                  & \multicolumn{1}{c|}{Not the learner}       & Not the learner            \\
\textbf{Data recorded}           & \multicolumn{1}{c|}{Learner's internal state} & \multicolumn{1}{c|}{Learner's internal state} & Learner's internal state & \multicolumn{1}{c|}{Visual}                & Sensor            
\end{tabular}
}
\end{table}

\subsection{Direct Demonstration}
Direct demonstration consists of techniques where the embodiment mapping is unnecessary because the learning agent performed the demonstrations. Hence, there is no need to convert the state-actions from the space of the demonstrator to the learner since it remains the same. However, there may be a need for record mapping if the state-action pairs performed by the demonstrator cannot be directly recorded in the data set.

\subsubsection{Teleoperation}

Teleoperation is the most direct method for transferring information about the behavior of the teacher to the learner. It is a setting where the human teacher operates the learning agent or an agent structurally identical to the learner. This agent can be a physical or simulated robot or a simulated agent, such as characters in video games. The state-action pairs of the demonstration are recorded directly from the learning agent's sensors. Because the agent that performs the task is structurally identical to the learner and the state-action pairs are extracted directly from the agent's sensors, there is no need for either type of mapping.

The main advantage of teleoperation is that it can be easily used in simulation environments and video games, unlike the other approaches. Furthermore, it facilitates data collection, which further eases the creation of new methods and benchmarks. The downside to teleoperation is that it requires the agent to have the ability to be manually controlled. Therefore, it cannot be applied to all problems. This control can be done through a wide range of interfaces such as joysticks and graphical or virtual-reality user interfaces. Another disadvantage is that not all users are able to teleoperate the agent to a degree that allows them to demonstrate the tasks, without extensive training. This requires the user to be technically proficient, denying one of the aforementioned advantages of demonstration learning.

Demonstrations using teleoperation have been applied to a wide variety of applications.
In \cite{helicopter} and \cite{NASA} data of a robot helicopter flight, using a joystick, was recorded and used to train an autonomous agent through reinforcement learning. For manipulation tasks, \cite{toiletpaper} trained a robotic arm from demonstrations to change rolls on a paper roll holder. 
A humanoid robot was teleoperated by a person, using Virtual Reality technology that transforms the operator's arm and hand motions into those of the robot, to create demonstration data and learn a manipulation policy from it \cite{NASA}. 
Teleoperation has also been used in simulated environments. In \cite{soccer}, a Robosoccer robot is trained by transferring the skills of humans playing Robosoccer to the robot through demonstration learning where the demonstrations were captured through teleoperation. 
In \cite{alien}, the authors teleoperated a PR2 robot to touch a red cube on top of a table to generate demonstration data. The data was then used to train the robot to associate movements to labels and be able to perform a sequence of trajectories from labeled data.

\subsubsection{Kinesthetic}
Kinesthetic teaching is the alternative to teleoperation where an external mechanism to control the agent is not available. Here the teacher physically manipulates the agent, moving the agent's joints through the correct positions that allow the agent to perform the task.
Alternatively, the movements can be controlled through speech, where the robot is told specifically what to do.
Similarly to teleoperation, the demonstration data is also captured from the agent's sensors, so there is no need for any mapping. The quality of the demonstrations depends on the capabilities of the human teacher. Even with expert demonstrators, the data obtained from kinesthetic teaching often requires the usage of post-processing techniques.

The applications of kinesthetic demonstrations are similar to the ones of teleoperation. However, kinesthetic demonstrations are restricted to physical agents and are mostly applied to manipulation tasks. A learning method for collaborative and assistive robots based on imitation learning using kinesthetic demonstrations was applied to a robotic arm in various assistive scenarios \cite{kinesthetic}. To extract features from each state for kinesthetic teaching, a system that captures desired behaviors in the joint space was created in \cite{jointspace}. In \cite{fewdemos}, demonstrations are captured through kinesthetic teaching to model the reward function for manipulation tasks.

\subsubsection{Shadowing}

Demonstrations performed through shadowing are performed by the learning agent. Since the demonstration data is captured through the agent's sensors, there is no need for embodiment mapping. However, the agent performs the task by copying the movements of the teacher through some form of tracking. There is a record mapping that converts the actions performed by the teacher to the actions of the learning agent.
 
Shadowing has been applied to humanoid robots and navigation. In \cite{ogino}, a humanoid robot learns to imitate a human demonstrator's arm gestures and is tested on a turn-taking gesture game. For navigation, a mobile robot learns routes demonstrated by a teacher in \cite{Nehmzow}.

\subsection{Indirect Demonstration}
Indirect demonstration consists of techniques where an embodiment mapping is necessary. The demonstration data is not captured directly from the learning agent's sensors because a different agent demonstrates the task. Hence, the agent can not directly apply the state-action pairs in the demonstration data set.

\subsubsection{Observation}

In the case of indirect demonstration by observation, a teacher performs the task and the demonstrations are recorded by sensors located externally from the learner. The sensor is often a camera or a set of cameras. The cameras are sometimes reinforced by additional sensors placed on the teacher. The agent is a passive observer during this process.
The advantage of this technique is its simplicity regarding the data collection process. The downside is that both types of mapping are required to convert the demonstrations into usable state-action pairs for the agent. Since the teacher's actual state-action pairs during the demonstration are not directly recorded, only image representations, there is a need for record mapping. Similarly, because the learning agent did not perform the task, the state-action pairs required for it to learn the task are also not available. Hence, there is also a need for embodiment mapping to map the data set's demonstrations into usable state-action pairs.
This is often done through machine learning techniques, usually through the extraction of features or automatic conversion of the teacher's context to the learner. 

Because this technique requires both types of mapping, even though it is the simplest form of demonstrating a task, the other techniques should be preferred when possible. Observation is more suitable for settings with high degrees of freedom where other demonstration techniques are more difficult to perform. Furthermore, errors linked to camera recordings, such as occlusions, blurriness, and noise, affect the recorded data and require extra post-processing steps. 

Demonstration learning through observation is the most applicable form of demonstration learning. It has been applied to manipulation tasks where a robot learns an assembly task from demonstrations in \cite{legos}, house chores both in real-world and simulation environment in \cite{gupta}, or to manipulate a piece of cloth and create different shapes with it \cite{knot}. Early works have a robotic arm learn to balance a pole from demonstrations \cite{pole}. In \cite{Dillmann}, a robot demonstrates a task and transfers its skill through demonstration learning to a different robot. Often demonstration learning from observation is combined with information sources other than cameras. For instance, a robot learns to grasp objects by learning from demonstration data captured from visual observations and a force-sensing glove \cite{tecnico}. 

\subsubsection{Sensors on Teacher}

Sensors on a teacher is a demonstration technique where the demonstrations are recorded from sensors on the teacher such as wearable devices \cite{wearable}. Therefore there is no need for record mapping. However, because the learning agent did not perform the task, the recorded data can not be directly used by the agent. This means that an embodiment mapping is required to convert the teacher's state-action pairs to the context of the learning agent.
Because of this requirement, these approaches are used when the learning agent can not be controlled through the task. Otherwise, teleoperation, kinesthetic, or shadowing techniques should be preferred. Human teachers are commonly used to demonstrate a task using their own bodies to train a humanoid robot. However, it still requires specialized sensors, such as human-wearable sensor suits, and the mapping mechanism to apply to the learner. 
The advantage of such techniques over passive observation is the precision it provides over the collected data due to the usage of sensors, unlike observation, where the teacher's actions must be inferred through record mapping. 

The information of sensors placed on a human is converted into the joint angles of a humanoid robot in \cite{speert} to learn to perform reaching and drawing movements with one arm and tennis swings. The approach has also been used to teach walking patterns to a biped robot in \cite{walking} to simulate human-like walking.
In \cite{customglove}, a custom glove was used to capture hand position and tactile information to record demonstration data performed directly by a human. The data was then used to obtain object model representations and optimize the policy to perform the task.

\subsection{Data Representation}

The demonstration data needs to be recorded in a structured way. 
The structure is dependent on what technique was used for recording the demonstration and will determine which algorithms can be used to train the policy. The state representation is called a feature vector. The feature vectors may include different information types, such as information about the agent's state, the environment, individual objects, etc. Particularly, the environment is often inadequate to be represented in its entirety because of its high dimensionality. Additionally, it often contains redundant and irrelevant information to the learning task. Hence, the selection of features needs to be adequate and efficient to convey enough relevant information to estimate a quality policy. The actions performed by the teacher are also normally included in the demonstrations. However, some approaches overcome the unavailability of actions in the data set and learn to infer the action that caused a state transition \cite{GAIL}.
Additionally, if the goal is to maximize a reward function, the rewards for each state-action transition can also be included in the data set. Lastly, extra information can be included in the data set such as indications of episode termination, and safety constraint violations, among others \cite{constraint,intervention}.

\subsubsection{Raw Features}

Raw features are features received from sensors during the demonstration and directly stored in the data set.
Hence, used in cases where there is no record mapping. 
The features may inherit noise from the sensors and may need to be pre-processed. If the features contain all the information required to learning the task, they can be directly used for training, and no further processing is necessary.

Such features can be easily obtained in virtual environments such as simulators or video games. In \cite{deepq}, the agent learns to play a set of Atari games, where the state-action observations are direct screenshots of the game. However, in the real world, these features are often unavailable and require extra sensors.

\subsubsection{Manually Designed Features}

Manually designed features are extracted using specifically designed functions. These functions were carefully crafted after analyzing the problem at hand. They convert the information in the raw data provided by the sensors used to record the demonstration into a more efficient structure that can be used to train the agent. These functions convert the information into a different structure and filter the irrelevant and redundant information, often reducing the dimensionality.

In \cite{mataric}, a robot is trained to imitate human movement from observations. The authors define key points on the human demonstrator, and the agent learns to detect motions by identifying changes in the key points' locations. In \cite{demiris}, object positions are tracked and used as features of the demonstrations. For video games, \cite{mario} obtains screenshots of the Mario Bros game and divides them into binary cells, each signifying if the respective cell contains objects.  

\subsubsection{Extracted Features}

Extracted features that are obtained from a learned function. These functions serve the same purpose as above: to process the raw data obtained from the sensors during the demonstration and identify relevant information from the raw data that can be used to learn the policy. However, these functions are not handcrafted by experts after carefully analyzing the problem. These functions are obtained through a model specifically trained for extracting features from the raw data. The model is typically a neural network and works as a black box, extracting the features in a way unknown to the programmer. These models are used when task-specific features can not be identified after an expert evaluation. Therefore, automatic feature extraction models have the advantage of minimizing the task-specific knowledge required. Additionally, these models have broader applicability and are not restricted to a specific task because they can be trained to extract features from any task. Therefore, they allow for the creation of a more general demonstration learning pipeline.

\cite{guo} uses deep learning to extract features for training an agent to play a set of Atari games. In \cite{levine}, features are automatically extracted from observations to train a robot to perform various manipulation tasks. In \cite{sampleeff}, an encoder extracts features from the state observations. Here, the encoder is trained simultaneously with the policy in an off-policy manner, improving sample efficiency.

\subsubsection{Time as a Feature}

Here, the demonstrations consist of time-action pairs. Such features can be applied to tasks where it can be assumed that the environment's state depends solely on time. Hence the agent can learn the ideal action for a specific time interval, and its behavior will always be optimal. It is assumed that there is a time loop synchronized with a fixed sequence of environmental states.
The overhead of designing such features and algorithms is reduced tremendously. Furthermore, the efficiency of such policies will be high while the aforementioned time synchronization requirements remain the same. The main restriction of such features is their applicability, since rarely does the state of the environment depend solely on time.

Other limitations of time-based policies are the robustness and dependency on the task. Since the policy depends heavily on the time, it is very susceptible to any changes in the environment that can invalidate synchronization. 
Moreover, the time-action pairs are specific to the task; such restrictions are difficult to scale or adapt even to a similar task. In \cite{tcn, atnet}, the different demonstration sequences are synchronized. Frames from different demonstration sequences with the same sequence index show the same behavior (represent the same point of the task). The model is then trained to extract features from the frames. The model aims to generate similar features for frames with the same timestamp and different features for frames with different timestamps.

\subsection{Data set Limitations}
\label{datalimit}

The behavior of the policy is directly dependent on the information provided by the demonstrations in the data set. In this section, we discuss the many ways the limitations of the data set can hinder the performance of the agent, and summarize important properties to take into account when designing or using a demonstration data set.

\subsubsection{Incomplete Data}
\label{incomplete}

The demonstration data set will represent a distribution that is a subset of the full MDP. The larger the distribution sample, the easier it is to generalize and tackle the curse of dimensionality problem. Furthermore, the less likely it is to encounter out-of-distribution states and have to deal with the problem of distributional shift.
However, real-world demonstrations are difficult to collect such that they adequately cover the MDP. Demonstrating every possible state-action pair is challenging and, in some situations, impossible such as in continuous spaces. If a specific state is missing from the demonstrations, the learner cannot estimate its best action during policy estimation. This section presents the ways demonstration learning approaches tackle missing data points in the data set. Therefore, human demonstrations likely generate narrow distributions. In such cases, it is critical to deal with the distributional shift by avoiding exiting the data set's distribution and encounter out-of-distribution states. Another type of incomplete data is the non-inclusion of rewards in the data set or the inclusion of sparse rewards. The creation of a reward function is often difficult. This is further complicated for complex state and action spaces. Hence, avoiding rewards or designing sparse rewards is easier. However, the less the feedback the harder the problem becomes. 

The simplest idea to deal with limited data is to obtain new demonstrations. In such approaches, as the learner interacts with the system, it encounters unseen states and requests a demonstration from the teacher for that given state. 
In \cite{veloso}, the authors introduce a confident execution approach, which focuses on learning relevant parts of the task where the agent identifies the need to request demonstrations. The agent selects between demonstration and autonomous execution. As the agent learns the task, it increases its autonomy, reducing both the teacher's training time and workload. \cite{dogged} tackles this problem by having the agent show its confidence score for performing its chosen action to the teacher. The teacher can choose to intervene and demonstrate or accept the action \cite{smile,intervention}.

The previous approach requires extra overhead to identify which states are missing from the data set and extra commitment from the teacher during the learning stage. The alternative approach corresponds to generalizing using the available data. One way to generalize is to create new data from the existing set.
Data augmentation is often used in machine learning to enlarge the data set and improve generalization to unseen data. Such techniques can be applied to state representations to generate unseen data points. In \cite{s4rl,rad}, different data augmentation schemes are compared and applied to off-the-shelf RL algorithms. However, naively applying data augmentation to DL/RL can cause new problems. The authors of \cite{svea} identify pitfalls for naively applying transformations to RL algorithms and then teach how to properly use them. In \cite{sunrise}, the instability problem is tackled by estimating the Q-values from an ensemble of agents. Another approach is to perform stitching. Stitching is the process of combining portions of different unsuccessful trajectories to solve a task.

Another approach is to use transfer learning methods and learn from data of other tasks. In \cite{leverage}, it is shown that in certain conditions, the problem of learning from few demonstrations of a task can be mitigated by using demonstrations of other tasks. If the rewards are unusable for the host task or unavailable in the data set, these can be set to zero. Additionally, the performance can be improved by applying re-weighting methods to the transitions.

Alternatively, to generate new data without requiring extra effort from the teacher, the learner can interact with the environment. In such approaches, the learner is pre-trained on the available demonstration data and is then fine-tuned with reinforcement learning to collect the remaining data. The restrictions of reinforcement learning, such as safety during exploration, are lessened because the pre-trained agent is more competent than an agent being trained from scratch through reinforcement learning. This approach does require balancing the exploration of new data using and exploiting the available data during policy estimation. Furthermore, it requires creating an exploration policy and a reward function that gives feedback based on the agent's action at a given state.

In \cite{deepq}, the agent is pre-trained on demonstration data before interacting with the environment. Then, the agent's policy is updated using both the demonstration data and the exploration data.
In \cite{daap}, two policies are learned simultaneously. One executes the task, while the other ensures that the first does not violate constraints that could lead to harmful results. Another approach is created in \cite{rarl}. Here, a second agent is trained to make the learning of the main agent as difficult as possible. By creating difficult settings during learning, the resulting policy is more robust than it otherwise would have been. 

Some methods explore the state space by maximizing the entropy of the visited state distribution \cite{proto,planworld}. In \cite{proto}, a policy is trained to explore the state space while estimating the representations. The state space is clustered, and the exploration rewarded is proportional to the distance between the visited states and the nearest cluster. In \cite{planworld} a world model is estimated in conjunction with an exploration policy. The policy is rewarded for maximizing the variance of the predictions of an ensemble of networks. In \cite{safecont,cbf}, safe exploration is achieved by constraining the policy on states which respect the safe values of pre-defined restrictions.

\subsubsection{Inadequate Data}

Most demonstration learning approaches assume that the quality of the data in the demonstration data set is optimal. However, this often is not the case. The demonstrated behavior can be sub-optimal, which in some cases can be intended if the goal is for the policy's behavior to appear human. Additionally, the data can have noise inherited from the sensors, blurriness, and occlusions. The data can also be redundant and unevenly distributed. In \cite{causal}, the authors create two algorithms for dealing with demonstrations corrupted by noise. The algorithms extract the idea of the expert demonstrator using Instrumental Variable Regression techniques from econometrics. 

Another issue is ambiguity. Data ambiguity is when there is inconsistency in the teacher's choices, selecting different actions for the same state in different demonstrations. 
The result is that a single state is mapped to multiple different actions in the data set.

Additionally, the data may contain unsuccessful demonstrations. If these are labeled as unsuccessful, or contain the reward information which the policy aims to maximize, the policy can be more robust by learning not to perform the state-action pairs in such demonstrations. However, if they are treated as successful demonstrations, such as in behavior cloning, they hinder the quality of the resulting policy. In short, the quality of the learned policy is directly affected by the quality of the data, and many factors exist that can affect the quality.

Some approaches make the best of sub-optimal demonstrations and use them to generalize and obtain smoother behaviors. In \cite{aleotti}, repeated demonstrations are used to encourage such behavior and smooth the policy. In \cite{pook}, data from multiple teachers is used to difficult the training of the learning agent, resulting in a more robust policy. 
Some approaches identify inadequate demonstrations and choose to remove them from the data set before training the policy. \cite{kaiser} identifies the sources of inadequacy and presents ways of tackling it. 
In \cite{learningsoft}, the authors use both successful and failed demonstrations. They separate the two types of data into clusters using an adapted version of Gaussian Mixture Models. Then they use the Gaussian components from the cluster of successful demonstrations to perform regression and generate better trajectories.

Other solutions deal with inadequate data by asking the teacher for more demonstrations, as mentioned in section \ref{incomplete}.
Reinforcement learning can be used to deal with poor-quality data, by collecting new interaction data. The learner is pre-trained on the available data and then is fine-tuned with the data collected by exploration approaches. The learner interacts with the environment and fine-tunes its policy based on the given feedback. This feedback can be given through a standard reward function or by a teacher.
In \cite{farahmand} the sub-optimal demonstrations are only used to constraint a reinforcement learning algorithm and prevent the agent from committing errors during exploration that would lead to harmful consequences.
In \cite{grollman}, the authors realize that there is useful information in failed demonstrations. They propose a method that trains a policy from failed demonstrations. Here, the agent is trained to avoid repeating such unsuccessful behaviors. In \cite{provablyeff}, an algorithm is proposed for learning policies from partially observable state environments.
Alternatively, \cite{expertise} choose to estimate the quality of demonstrations by estimating the competence of the demonstrator and filtering the transitions based on the competence level.

\section{Learning from Demonstrations}
\label{policy}
In this section, we explain the different methods available in the literature for using the demonstration data set. In general, the demonstration learning methods learn a policy or a world model. However, the demonstration data sets have also been used to learn other types of models which we will discuss. 

\subsection{Learning Problems}

Demonstration learning relies on the data set to learn the models. As discussed in Section \ref{datalimit}, collecting the perfect data set is unfeasible for most applications and therefore it will contain some limitations which affect the learning method.
To counter the limited demonstrations, the methods should aim to generalize to regions outside the demonstrated regions in the data set. However, if the data set does not contain transitions that correspond to high-value decisions, such as high associated rewards, it may be impossible to discover those regions. \cite{levinesurvey} argues that there is nothing that can be done about this challenge and that methods should assume that the data set contains enough information for a suitable model. To counter imperfect demonstrations, the method should filter good demonstrations from bad ones. Moreover, the method should learn to extract the good parts of the demonstration, avoid the bad parts and potentially combine parts from multiple demonstrations. Naive imitation in a self-supervised manner, through behavior cloning, copies bad behaviors. Hence, methods that filter bad demonstrations, can obtain a better policy than the one represented by the data set.

Another problem with demonstration learning is that it is in its essence a paradox. Demonstration learning combines supervised learning with transition data of reinforcement learning. To improve upon the policy of the data set, the goal is to answer what are the sequences of actions that generate the maximum reward. However, supervised learning methods assume that the data is independent and identically distributed (i.i.d.). The model should obtain good performance as long as the data it encounters comes from the same distribution as the one it was trained on. However, in demonstration learning, the goal is often to mimic or improve upon the behavior observed in the data set.

All these problems could be alleviated by interacting with the environment and testing uncertainties that the method may have. Hence why demonstration learning is often followed by reinforcement learning for refining the policy with online interactions. Pure demonstration learning is difficult because the agent can not collect extra transitions and explore new regions.
Technically, any off-policy method existing for online reinforcement learning could be used to learn a model from the demonstration data set. However, these methods were created with the assumption that the agent could interact with the environment to perform corrections to existing errors.
Demonstration learning estimates a model to perform a task defined by the state-action distribution $\Delta_{task}$. Demonstration learning methods estimate the model using the demonstration data set which contains a set of transitions. The data set also has an associated distribution $\Delta_{demo} \subset \Delta_{task}$, which is a subset of the task distribution. However, during deployment, the agent will likely encounter regions outside the distribution of the data set, $(s,a) \notin \Delta_{demo}$. The prediction of the model for such regions will result in larger mistakes than for in-distribution regions. Furthermore, these mistakes will accumulate and the agent will continue to diverge from the learned distribution. This snowball effect is called the 'distributional shift'. Most policy learning methods through offline reinforcement learning tackle the distributional shift directly through different manners. 
Some use behavior cloning to restrict the distribution to the distribution of the data set. This heavily restricts generalization. Others propose to punish the distributional shift in the training loss by an estimation of uncertainty. Others constrain the agent to specific regions by making conservative estimates of future rewards for the Bellman update, by learning a lower bound estimation of the true value function.
In \cite{mario}, the authors proved that even with optimal action labels, the compound errors of distributional shift accumulate to a quadratic error in the best-case scenario. However, this error would scale linearly, if the agent was allowed to collect extra transitions.
Demonstration learning methods struggle to find a balance between generalization and avoiding the distributional shift.

\subsection{Policy Learning}

\begin{figure}[!t]%
    \centering
    \subfloat{{\includegraphics[width=0.25\textwidth]{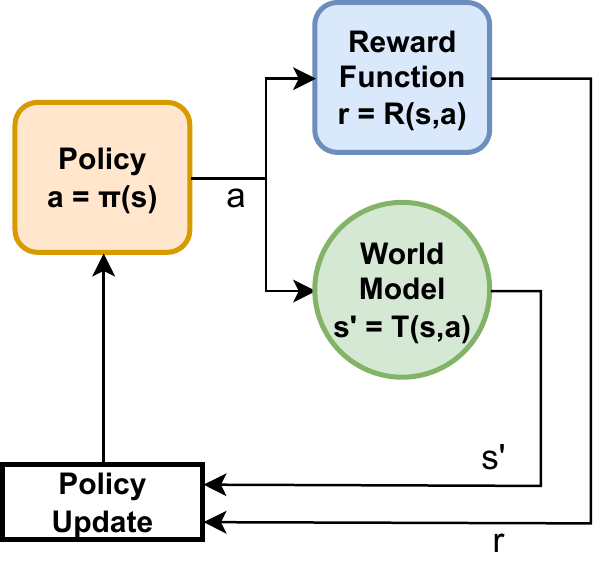} }}%
    \qquad
    \subfloat{{\includegraphics[width=0.25\textwidth]{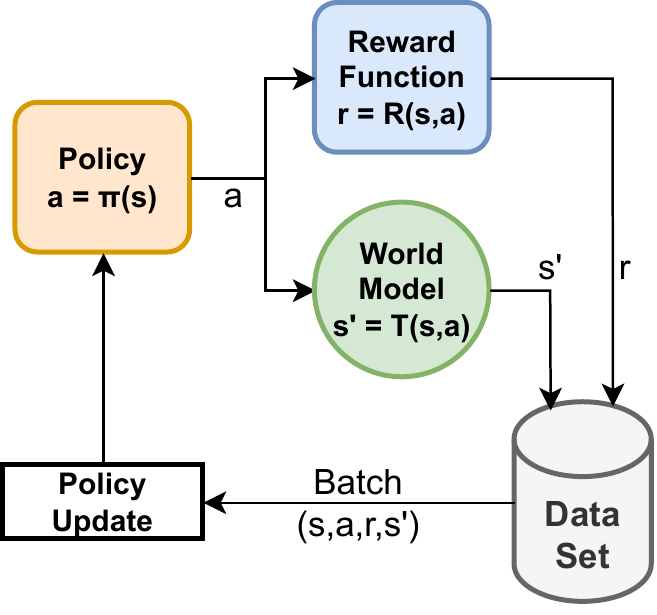} }}%
    \qquad
    \subfloat{{\includegraphics[width=0.25\textwidth]{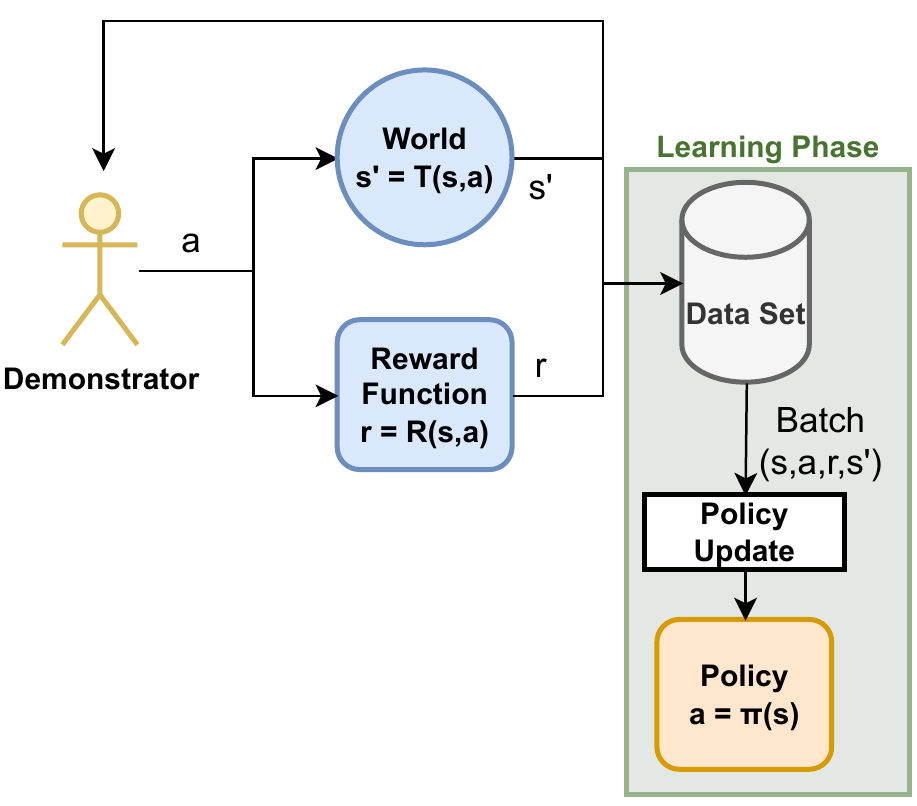} }}%
    \caption{Differences between on-policy reinforcement learning, off-policy reinforcement learning, and demonstration learning.}%
    \label{policyfig}
\end{figure}

Policy learning from demonstrations involves learning the correct mapping from states to actions from the demonstration data set. The teacher demonstrated a policy $\pi_{teacher}$ and the demonstrated state-action pairs in the data set are examples of the correct mapping. The closer the estimated mapping function is to the original mapping in the data set, the better the agent will reproduce the teacher's behavior. Reinforcement learning estimates the policy by interacting with the environment, receiving a reward, and adjusting the policy accordingly. Additionally, the agent can maintain a history of past interactions and use them to continuously update the policy. This setting is named off-policy reinforcement learning because the policy is being updated with data collected by a previous policy. In demonstration learning, the agent learns from recorded data from the start, through the demonstration data set. The differences between the three settings are summarized in Fig. \ref{policyfig}.

\subsubsection{Behavior Cloning}

Behavior cloning is the simplest form of obtaining a policy from a demonstration data set. Here the policy is trained to directly imitate the teacher's action for all the states in the data set. The problem corresponds to either a classification or regression problem, for discrete or continuous action spaces, respectively. Formally, the policy is trained to minimize the error between its predicted action and the ground truth action for all state-action pairs in the data set:

$\pi_\theta* = \arg\min_\theta (\pi_\theta(s) - a), \forall (s,a) \in D_{demo}$.

Another approach is to maximize the likelihood of actions in the demonstration:

$\max\mathop{\mathbb{E}}_{(s,a)\sim \mathop{\mathbb{D}}}  log \pi(a|s)$. 

However, because behavior cloning is a naive copy of the data set, it is more reliant on the quality and size of the demonstration data set than other alternatives. The data set corresponds to a sub-set distribution $\Delta_{demo}(s|a)$ of the real distribution of states over actions for a given task $\Delta(s|a)$. Behavior cloning gives guarantees that the agent performs so long as it only encounters states present in the demonstration data set. However, no such guarantees are present if the agent should encounter an unseen state. In \cite{expertise}, the authors tackle the susceptibility of behavior cloning to the quality of demonstrations by estimating the competence of the demonstrator and filtering the transitions based on the competence level.

\subsubsection{Offline Reinforcement Learning}

Sometimes, direct imitation through behavior cloning is not adequate to reproduce the behavior and solve the task due to errors in the demonstration or poor generalization.
The term offline reinforcement learning has been used to define many methods and is sometimes used as a synonym for demonstration learning. Here, we choose to use this nomenclature to aggregate methods that employ reinforcement learning techniques to a data set of demonstrations.

In offline reinforcement learning, the agent has access to the rewards attributed by the environment to each transition. The policy is trained to maximize the expected accumulated reward $J(\pi) = \mathbb{E}[\sum_{t=0}^{\infty} \gamma R(s_t, a_t)]$, where $\gamma$ is the discount factor.

In general, all reinforcement learning algorithms follow the same
basic train loop. The agent observes the current environment state $s \in S$, then it interacts with the environment by selecting an action from its policy $a_t \sim \pi(s_t)$, the interaction changes the state of the environment to $s_{t+1}$ and the agent receives a reward $r_t = R(s_t,a_t)$. This repeats for multiple interactions. The agent stores the transitions $(s_t, a_t, s_{t+1}, r_t)$ in its memory and uses them to update the policy. In offline reinforcement learning, the memory is given by the demonstration data set $D_{demo}$.

Due to the limitations of behavior cloning, some approaches choose to pre-train the agent on demonstration data. Then, optimize the agent to learn the remaining state-action space by employing online reinforcement learning \cite{deepq}. 
However, online reinforcement learning is dangerous as some actions can lead the agent to catastrophic states which are unrecoverable in real-world scenarios. Because of this, some approaches choose to employ pure offline reinforcement learning and apply regulators to reduce the impacts of distributional shift \cite{cql} or prevent the agent from going out of distribution \cite{ldm}.

One way to optimize the policy, parameterized by weights $\theta$, for the Bellman objective is to estimate the gradient:

$\nabla_\theta J(\pi_\theta) = \mathbb{E}_{(s_t,a_t,s_{t+1},r_t) \in D_{demo}}[\sum_{t=0}^{H} \gamma
^t \nabla_\theta \log \pi_\theta(a_t \mid s_t) Q^\pi(s_t,a_t)]$, where $Q^\pi(s_t,a_t)$ is the state-action value function.

Alternatively, we can use dynamic programming methods by first estimating the state or state-action value functions, and then using them to optimize the policy. The state value function $V^\pi(s_t)$ returns the estimated expected accumulated reward that can be obtained by starting at state $s_t$:
$V^\pi(s_t) = \mathbb{E}_{(s_t,a_t,s_{t+1},r_t) \in D_{demo}}[\sum_{t'=t}^{H} \gamma^{t'-t} R(s_t,a_t)]$

The state-action value function $Q^pi(s_t, a_t)$ is similar and returns an estimation of the expected accumulated reward that can be obtained by starting at state $s_t$ and performing action $a_t$:

$Q^\pi(s_t,a_t) = \mathbb{E}_{(s_t,a_t,s_{t+1},r_t) \in D_{demo}}[\sum_{t'=t}^{H}\gamma^{t'-t} R(s_t,a_t)]$.

From these definitions, we can reformulate them into a recursive form:
$Q^\pi(s_t, a_t) = R(s_t,a_t) + \gamma \mathbb{E}_{s_{t+1} \sim P(s_{t+1}\mid s_t, a_t), a_{t+1} \sim \pi(a_{t+1}\mid s_{t+1})} [Q^\pi(s_{t+1}, a_{t+1})]$.

The algorithms that estimate the policy based on dynamic programming are mainly split into two families: Q-learning and Actor-Critic methods. 

In Q-learning, the policy is obtained directly by estimating the state-action value function, and selecting the action that maximizes the expected accumulated reward: $\pi(a_t\mid s_t) = \arg \max_{a_t} Q(s_t, a_t)$. The Q-learning objective is defined by $Q_\theta(a_t,s_t) = R(s_t,a_t) + \gamma \mathbb{E}_{(s_{t+1}\mid s_t, a_t)} [\max_{a_{t+1}}Q_\theta(s_{t+1},a{t+1})]$.

Actor-critic algorithms are a mixture of policy gradients and dynamic programming because they use a policy, the actor, like policy gradients, but also use a value function, the critic, like dynamic programming. Actor-critic algorithms learn the state-action value for the current policy $\pi_\theta(s_t)$:
$Q^\pi(s_t,a_t) = R(s_t,a_t) + \gamma \mathbb{E}_{s_{t+1} \sim P(s_{t+1}\mid s_t, a_t), a_{t+1} \mid \pi(s_{t+1})}[Q^\pi(s_{t+1},a_{t+1})]$.

In early research, a set of algorithms were explored to fasten and improve reinforcement learning.  The authors in \cite{lin}, compared eight reinforcement learning frameworks, including pre-training with demonstration data, for performing a task of playing a 2D game. The authors concluded that pre-training the policy on demonstration data prevents the learner from falling in a local minimum and increases its scores. Furthermore, the improvements are more noticeable with the increase in the difficulty of the task. For more difficult tasks, pre-training on demonstration data significantly increases performance. 
Pre-training a reinforcement-learning policy using demonstrations has resulted in the creation of an agent capable of playing the game "Go" in a way that rivals the best human players \cite{go}. 
In these applications, the initial policy's weights are obtained from training in demonstration data. Then, the weights are updated from exploration using reinforcement learning.
In \cite{deepq}, the authors tackled the issue of reinforcement learning not being applicable to real-world issues, due to learning from trial and error and errors in the real-world having serious consequences. They pre-trained the policy on demonstration data, and the results showed higher rewards for the first learning iterations when compared to standard reinforcement learning approaches. Therefore, pre-training on demonstration data results in safer exploration.

Additionally, a policy trained from online interactions can then demonstrate correct interactions. These demonstrations can then be used to transfer the knowledge to another agent. This approach does not need a human teacher because the policy is learned from scratch using trial and error, and then the interactions are recorded as demonstrations. This can be useful in cases where the policy updates can't be performed in real-time \cite{guo}.

\subsubsection{Classification}

Classification, in this context, consists of attributing the action class to the input state. Such techniques are employed when the action domain is discrete and defined as a set of finite and specific individual actions. 
%The policy's input represents the world and/or the agent, the same way it was represented in the data set. The output corresponds to a single action of the set of possible actions the agent can perform. The problem determines the set of actions. 
For example, in a 2D platform game where the agent can only walk left or right or jump, the inputs are categorized into these three actions. The policy's performance is evaluated by how often it attributes the correct action for any given input state.

Formally, the policy is a classifier $\pi(s)$, used to predict the action class $a$ of an observation $s$. Where $a \in A, A = \{a_1, ... a_n\}$ is a finite set of actions.

Classification methods can be applied to different levels of complexity ranging from low-level actions to complex behaviors. Bayesian networks were used in \cite{inamura} for navigating an environment and avoiding obstacles.
In \cite{saunders}, the authors created mapped representations of the environment and selected the robot's actions using a k-nearest neighbors algorithm. In \cite{veloso}, a Gaussian mixture model (GMM) is used for classifying actions in navigational problems. In \cite{raza}, four classifiers were tested for robot soccer cooperation tasks.

More recently, neural networks have been the go-to classifiers. This is due to their applicability because they are universal approximators. Recurrent neural networks (RNN) are used in \cite{rahmatizadeh}. Here a robotic arm is trained from demonstration data collected in a simulation environment to perform manipulation tasks. The trajectory is learned using the RNN. The trajectories are obtained in real-time and consider the current position of the end-effector and the objects. Classifiers are often directly applicable to video games because there is a finite number of possible actions. In \cite{deepq}, the policy is represented by a neural network classifier with 18 neurons in the last layer, one per action.

Some works have proposed ways of discretizing continuous action space \cite{continuouscontroldiscrete}. This allows any discrete RL algorithm to be applied to the continuous state problem.

\subsubsection{Regression}

Regression, in this context, consists of selecting a set of scalar values that compose the action, given an input state. Such techniques are employed when the action domain is continuous. For example, in the control of a robotic arm, each action can be defined by the robot's joint angles. The policy's performance is evaluated by comparing the estimated action values with the ground-truth action values in the data set for the given state.

Formally, the policy is a regressor $\pi(s)$ which maps a state $s \in S$ to actions $a \in A$, where each action is defined by a finite set of continuous values $a = \{a_1,...,a_n | a_k \in \mathbb{R}\}$.
Typically, regression approaches are applied to low-level motions and not high-level behaviors because high-level behaviors are a combination of low-level motions and are more likely to be discretized.

A traditional regression technique is Locally Weighted Regression (LWR). It is suitable for learning trajectories, as these are made up of sequences of continuous values. In \cite{kober}, a robotic arm is trained to execute a trajectory that allows it to perform manipulation tasks. Locally Weighted Projection Regression (LWPR) extends the previous approach to cope and scale with the input data's dimensionality and redundancy. \cite{dogged} uses LWPR to have a robot perform basic soccer actions.

Similarly to classification, the most common paradigm of recent works is neural networks because they can represent any function. In \cite{tcn, atnet}, a robotic arm is trained from demonstrations to perform manipulation tasks. The action is given by a neural network where the number and values of the last layer's neurons correspond to the number of joints of the robot.

Other approaches choose to specify the type of task that the agent can perform in the network. In \cite{crosscontext}, a robotic arm is trained to pick and place blocks. The network which outputs the action has 4 neurons. The first 2 specify the position and rotation of the end-effector for picking a block while the other two are for placing a block. Therefore, the two groups of neurons are used separately.

In \cite{pseudometric}, an algorithm is proposed to convert discrete actions in the demonstration data set into continuous ones.
An encoder is trained for this purpose to promote behavioral and data-distributional relations in the features. 
Then, an off-the-shelf algorithm can be used to train a policy using the new data set. However, because the policy outputs feature embeddings, the actions can't be directly applied to the task environment. Therefore, the output of the policy is discretized by finding the action whose embedding is closer to the output of the policy. 

\subsection{Model Learning}

\begin{figure}[!t]%
    \centering
    \subfloat{{\includegraphics[width=0.35\textwidth]{imagens/demolearning.pdf} }}%
    \qquad
    \subfloat{{\includegraphics[width=0.35\textwidth]{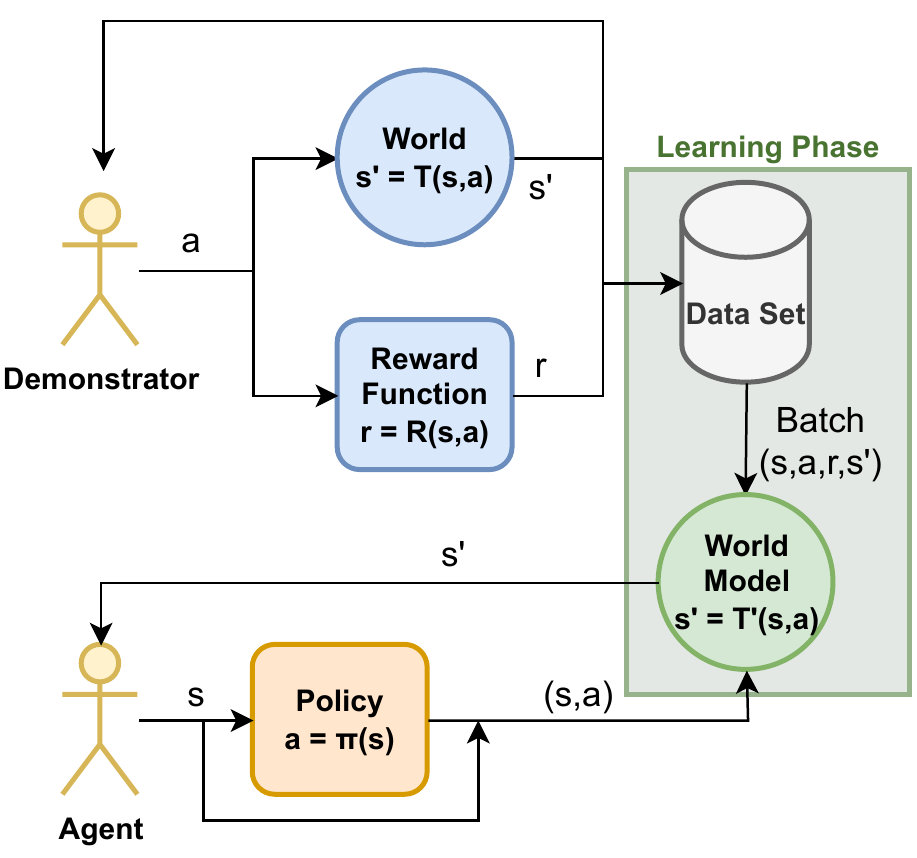} }}%
    \caption{Differences between policy learning and model learning from demonstrations.}%
    \label{modelfig}
\end{figure}

Model-based methods learn the dynamics of the environment by estimating the transition function $\psi(s_t, a_t) \sim P(s_{t+1}\mid s_t, a_t)$. The estimated transition function can be used as proxies of the real environment. Hence, the agent can collect new transitions without actually interacting with the environment, remaining safe during the learning process. In standard RL, the agent must first interact with the environment to collect transition data that represent the dynamics. In demonstration learning, the transition function can be estimated from the demonstration data set. These differences are represented in Fig. \ref{modelfig}. The functions are typically estimated through standard supervised regression using the states and actions as inputs and the next states as the desired output: $\mathop{\mathbb{L}}_\psi (s_t, a_t, s_{t+1}) = \|s_{t+1} - \psi(s_t, a_t)\|$. Model-based learning methods from standard RL can be used to learn from demonstrations \cite{janner,planworld}.
Standard online learning algorithms can be applied naively, with minimal modification, to train a model from demonstrated data.

However, because the policy learns from transitions simulated by the model, the performance of the policy is dependent on the quality of the estimated model which in turn is dependent on the quality and coverage of the distribution present in the data set.
In standard RL, the models can correct mistakes in the estimations by collecting new transitions. Similarly to policy learning, if the model is estimated solely from demonstrations, it can suffer from the distributional shift problem. 
In fact, the model can suffer from distribution shift regarding the true state distribution, and the true action distribution. 

The distributional shift can cause the model to be exploited by the policy. The policy is optimized to maximize the expected accumulated rewards. The policy can use the model to produce out-of-distribution states. Because these states are out-of-distribution the predicted values of the model are likely incorrect and may have an associated higher reward than the true state in the real MDP. Hence, the policy is learning to maximize erroneous transitions which lead to a worse performance once deployed to the real MDP.

A theoretical analysis is present in \cite{janner}, where they formulate the bounds on the error between the learned policy and the policy in the data set, due to distributional shifts in the policy and model.
The methods for reducing the distributional shift in policy learning can also be applied to model learning.
The main way to reduce the distributional shift problem in model learning is by learning an auxiliary model $\mathbb{U}(s,a): \mathbb{S}\times\mathbb{A}\rightarrow \mathbb{R}$, that punishes the reward function such that the agent avoids states outside the distribution: $R'(s,a) = R(s,a) + \mathbb{U}(s,a)$. Model-learning methods usually measure the uncertainty using an ensemble of models. In \cite{kidambi}, the method only punishes the reward function if the disagreement of the ensemble is above a threshold. Alternatively, \cite{mopo} uses a pessimistic approach by selecting the maximum prediction uncertainty of the ensemble. In both cases, the policy is penalized for visiting states where the model is likely to be incorrect.
Measuring uncertainty is challenging and often unreliable. In \cite{bremen}, the policy is learned by generating a new data set from rollouts on each of the models of an ensemble to counter uncertainty. Another approach to reduce the distributional shift without quantifying uncertainty is to use a regularizing term. In \cite{combo} a model-based version of the CQL \cite{cql} algorithm is proposed.

Instead of learning the policy inside the model, the learned model can be used to provide an evaluation of the learned policy without interacting with the environment. In \cite{farajtabar,brunskill,wang2017} the model provides an estimate of the expected return of the trajectories produced by the policy. In \cite{rafailov}, a model is learned from demonstrations to estimate a task from images, which due to their high dimensionality, is especially difficult.

\subsection{Inverse Reinforcement Learning}

\begin{figure}[!t]%
    \centering
    \subfloat{{\includegraphics[width=0.35\textwidth]{imagens/demolearning.pdf} }}%
    \qquad
    \subfloat{{\includegraphics[width=0.35\textwidth]{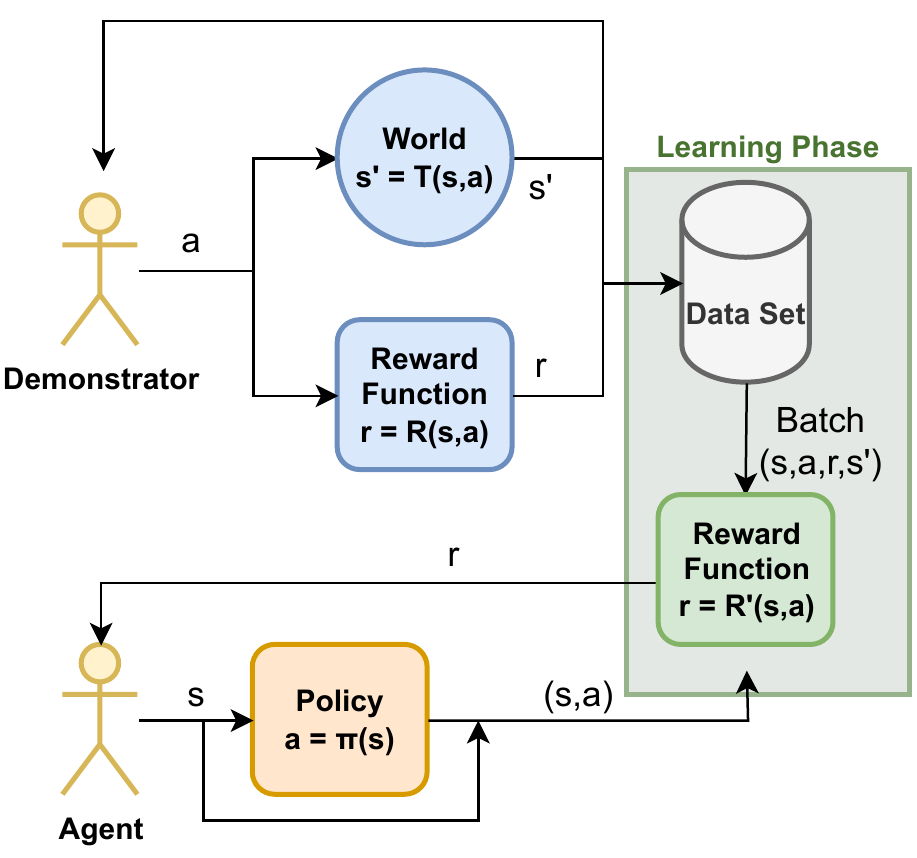} }}%
    \caption{Differences between policy learning and reward learning from demonstrations.}%
    \label{irlfig}
\end{figure}

Reward functions map a state transition to a reward value based on the quality of the interaction: $R(s,a): S \times A \rightarrow \mathbb{R}$. Reward functions determine what is desired from the agent in the task. Hence, they are used to guide the learning process of the agent, by estimating parameters that maximize the expected accumulated reward.
Usually, the reward functions are handcrafted by the programmer. This requires creating a function to map each state-action pair to a reward value. However, covering the entire state space is difficult for high-dimensional domains. Because of this, these functions tend to result in sparse rewards. Transitions where the agent receives no feedback, through a reward of zero, hinder the convergence of the policy to an optimal one and sometimes may prevent convergence altogether. The requirement to create a reward function that covers the entire task, limits the applicability of learning algorithms to problems where a reward function can be easily specified.

An alternative is inverse reinforcement learning IRL \cite{norvig}, also named reward shaping. Here, the demonstration data set is used to create a reward function. Then, a policy is estimated with off-the-shelf online reinforcement learning methods to maximize the expected accumulated reward defined by this function.
Hence, IRL offers a way to broaden the applicability of task learning models and reduce the manual work required by the programmers when demonstrations of the task are available. The differences between policy learning and reward learning from demonstrations are represented in Fig. \ref{irlfig}.
\cite{russel} points out that the reward function is more transferable than a policy. Even minor changes to the task likely render the policy useless. Such changes do not impact the reward
function nearly as much. Often, the learned reward function simply needs to be extended to new states.

Demonstration learning assumes that the teacher follows a policy $\pi_{teacher}$ which is maximizing a reward function $R_{teacher}(s,a)$ when demonstrating a skill. 
The idea of inverse reinforcement learning is to estimate the underlying reward function from the demonstrations.
Formally, we have an MDP without the reward function, $MDP\setminus_R$, and a demonstration data set with $N$ demonstrated trajectories $D_{demo} = \{\tau_i\}^N$, where each trajectory is a sequence of $L$ state-action pairs $\tau_i = \{(s_j,a_j)\}^L$. The goal is to create an estimate $\hat{R}$ the reward function that best describes the demonstrated behavior. Hence, IRL inverts the RL problem. Instead of learning an optimal policy from demonstrations, potentially using the logged reward $(s,a,r)$, IRL seeks to explain the demonstrated behavior by estimating the corresponding reward function.

IRL should estimate a reward function that generalizes from the demonstrated behavior. Hence, like other demonstration learning methods, aims at answering the question: what happens if the agent were to perform a trajectory different from the ones seen in the data set? This is important because if we want the learning agent to improve upon the behavior seen in the data set, the agent must execute a trajectory that is different than the ones in the data set. However, most machine learning algorithms assume that the data is independent and identically distributed (i.i.d.). Answering this question is difficult due to the problem of distributional shift.

Additionally, there can be many solutions to the reward function that describe the same behavior resulting in ambiguity. Moreover, some of these solutions can describe the problem while being unusable such as a function that always returns the same reward. Due to this ambiguity, it is important to determine how to measure the performance of the estimated reward function. If the true reward function is available, then we can measure the errors of the reward predictions for each demonstrated state-action pair with the ground truth reward. Alternatively, we can estimate a value function from the learned reward function and compare it to the real value function. However, the true reward function is often not available, and this unavailability is the reason to use IRL. A more general way of measuring performance is to estimate a policy from the learned reward function and then measure the performance of the policy using the demonstration data set. The limitation of this method is the problem of how to evaluate the policy. Interacting with the environment is not possible because the true reward function is not available. Hence the only policy evaluation metric is to compare the policy predictions with the actions of the demonstrator for every state in the data set. However, this comparison is limited because even if the policy is only wrong in a single state, it can still result in compound errors at deployment. No single metric in IRL gives a satisfying evaluation of performance. 
Additionally, the base structure of the function for which its parameters are estimated is non-trivial. Choosing too many parameters can lead to overfitting and prevent generalization. On the other hand, using too few parameters may prevent the policy from converging. 

To obtain a unique reward function, IRL methods define additional optimization goals, most commonly, maximum margin and maximum entropy. In the maximum margin setting, the reward function is the one that maximizes the difference between the best policy and all other policies. 
\cite{terrain} uses maximum-margin-based IRL to find a policy for performing navigation on rough terrain.
Contrarily, maximum-entropy finds a distribution of policies that maximizes the entropy with respect to specific constraints, such as feature matching, which ensures that the goal of the task is reached. \cite{ziebart} uses maximum entropy to learn a reward function for a driving task where there are multiple routes for the same goal in the demonstrations. The approach is later expanded to use deep learning in \cite{tennis} in a table tennis task. 

For discrete action spaces, IRL may be formulated as a classification problem, where for each state-action pair, the action is seen as the label for the state. The direct way to obtain a reward function is by estimating the action-value function which we explain in Section \ref{problem}. This approach was used by \cite{scirl} and later by \cite{csi}. However, such an approach relies on the principle that the demonstrated pairs are optimal.

Another approach to estimate the reward function is to associate states encountered in the demonstrations, or similar states, with higher rewards than the states not found in the data set. In \cite{fewdemos}, demonstrations are used to estimate a Hidden-Markov-Model which determines the associated reward for each state.
In \cite{pole} this approach is used in balancing a pole using a robotic arm.  In \cite{fb}, the authors experimented with three variants of reward function parameterizations from demonstrations and applied them to reaching, picking, and placing tasks. In \cite{tcn, atnet}, the reward is proportional to how close the images captured by the learning agent at a certain timestamp are to the respective frame of the demonstration video.

Some approaches choose to adopt an actor-critic algorithm, where the reward function is defined by a third-party critic who gives feedback based on the actions of the agent (actor). In \cite{daap}, a critic's policy is trained simultaneously with the agent's. Its goal is to prevent the actor from violating constraints pre-defined (engineered) beforehand that would lead to serious consequences. As the number of learning iterations grows, the actor becomes more competent, and the power of selecting the action increasingly shifts from the critic to the actor.

Such reward functions can encourage sub-goals or milestones during the task execution that are represented in the demonstrations.
In \cite{shaping,shaping2}, the authors investigated the intersection between reinforcement learning and demonstration learning. Results show that the approach based on reward-shaping can be more sample-efficient and more robust against sub-optimal and inconsistent demonstrations in two simulated domains than transfer learning algorithms.

\subsection{Other Learning Methods}

In this section, we discuss methods that complement or refine the previous learning methods to be more accurate, general, or robust.
Learning from demonstrations may not be enough to learn the task for all scenarios due to the limitations of the data set discussed in Section \ref{datalimit}. Interacting with the environment allows the agent to collect extra data that it may use to refine the model. 

The data set likely does not include a demonstration for all the possible environment states, this is especially true in high-dimensional spaces such as real-world tasks. 
Hence, the agent should aim to generalize during the learning process when using the data set. However, the agent may still not be able to generalize due to either limitation of the data set or the learning method.
Generalizing in demonstration learning is especially hard because the demonstrations are sequences of interactions. Because each interaction depends on the history of previous interactions, it violates the i.i.d. assumption of supervised learning to generalize \cite{smile}. 
During demonstration learning, the agent learned a sub-set distribution of the real task distribution. At inference time, if the agent encounters out-of-distribution states, it may not know what to do and the predicted action can be dangerous. As discussed previously, some methods try to reduce this issue by explicitly reducing the distributional shift. However, constraining the agent to states inside the distribution of the data set may limit the performance of the agent. 
Because of this, methods that refine the agent's model by interacting with the environment, allow the agent to learn missing information. Such an agent is safer than a random agent with no task knowledge performing random interactions.

\subsubsection{Reinforcement Learning}

RL models the problem as an MDP, as does demonstration learning. Instead of the agent learning from a data set of previous 
environment interactions, the agent will interact with the environment using its current policy and obtain the interaction rewards. RL starts with a random policy and tunes its parameters toward maximizing the expected accumulated rewards. However, the same idea can be used to refine the parameters of a policy learned from demonstrations. The agent initialized with a policy learned from demonstrations is safer than one with a random policy. Furthermore, the agent initialized with a policy learned from demonstrations converges much faster and avoids the risk of converging to a local minimum. These benefits were shown in learning atari games in \cite{deepq}. The agent performs exploration and learns unseen regions of the state space which increases its generalization capabilities and robustness. However, if the agent encounters unknown state regions, it may not know the correct action and will likely make a mistake. Even though it learns from the mistakes, in the real-world, the mistakes can have catastrophic consequences. Hence, such algorithms should be equipped with safety mechanisms. Reinforcement learning can also be used to train a policy from scratch in an online manner. This policy can then be used to generate the demonstration data set to train and evaluate the demonstration learning methods. This approach is limited to environments where online RL is possible. It is also mainly used to automatically generate demonstrations to evaluate the performance of demonstration learning methods. Because if online RL is possible and available there is no need to train a second policy with demonstration learning. However, in certain cases training a second policy can be advantageous if the RL agent does not act in real-time \cite{guo}.

Perhaps the most well-known application was training an agent to learn to play 'Go' to the extent of beating human experts \cite{go}. Here the agent is trained using demonstrations and then refined using RL. In \cite{zhang2016}, recurrent neural networks are used to deal with POMDPs by incorporating past information to guide decision-making. The agent is trained with RL while the demonstrations are used to decide which memories to store.

\subsubsection{Evolutionary Algorithms}

Like reinforcement learning, optimization algorithms can be used to learn or optimize a policy to replicate a behavior. Evolutionary algorithms (EA) are popular optimization methods, inspired by natural animal behaviors, used to find solutions to various problems.

EAs can be used to generate trajectories. The most common are Particle Swarm Optimization (PSO) \cite{zhang2015} 
and Ant Colony Optimization (ACO) \cite{zhang2010}.
The algorithms are inspired by the behavior of animals in this case, birds and ants, respectively, to find an optimal solution in the search space. They have been extended with demonstrations for improving the learning process.
In \cite{soccer}, EAs are used for optimizing agents trained in a soccer simulation. The possible solutions are represented by chromosomes and are a set of if-then rules. These were obtained from demonstration data. Then, the different solutions are evaluated by a function that measures their performance. The best-performing solutions survive to the next generation. 
In \cite{cheng}, PSO is used for finding the optimal behavior. The demonstrations define the initial behavior. Each particle modifies its behavior by observing better-performing particles. The performance of a particle is defined by a fitness function.
In \cite{bongard}, Preference based Policy Learning (PPL) is used to teach a robot to navigate.

\subsubsection{Transfer Learning}
Transfer Learning (TL) is a paradigm used to apply the knowledge acquired from training a task to learn a second task. Instead of training the second task from scratch, the knowledge of another task can be used as a starting point, optimization, or, in rare cases, to perform the task completely. Formally, given a task $T_s$ learned in the MDP domain $D_s$, the idea is to improve the learning of the goal task $T_g$ in the MDP domain $D_g$ using the knowledge of the previous task. Transfer learning is beneficial for reducing the need to gather new samples because they can either be directly used by the new task, or the knowledge gained from training a policy on the data can then be used to train the new task.

In demonstration learning, the demonstrations recorded for one task can be used to learn a second task. 
\cite{brys} use transfer learning to extend reward shaping. Reward shaping relies on prior knowledge. Therefore, transfer learning can use the knowledge of a policy learned for one task to perform reward shaping for a similar task.
\cite{kuhlmann} found that even if the agent overfits on the previous task, it can still adjust its weights enough to recover and converge to the ideal weights of the second task. Their approach is based on using graphs for identifying previously encountered games and using the relevant knowledge on the current game.

Alternatively, policies learned for a task can advise a learner on another task that shares similarities. The knowledge can be in the form of useful feature representations and specific parameter values. Additionally, a policy can be used as a starting point to learn a new policy for a different task.
In \cite{torrey}, transfer learning is used to learn a new soccer skill after having learned a different one in a simulation. 
Experiments show that transfer learning reduces the convergence time and achieves a better performance.

\subsubsection{Adaptive Learning}

Demonstration learning algorithms must take into account that the demonstration data set is likely incomplete and missing many regions of the state space. The most common way to tackle such problems is to use pessimistic or conservative methods, which keep the policy close to the regions of the data set and avoid behaviors that are too different. However, such approaches can recover sub-optimal estimations due to the strong restrictions. 
Additionally, agents can get stuck in certain states and repeat the same action over and over. In such cases, policies should be able to adapt to bad choices by applying corrections. In \cite{adaptivepolicies}, the authors train uncertainty-adaptive policies that receive a belief as an extra parameter. This belief is estimated from the history of interactions using an ensemble of networks. After a failed interaction, the history changes, causing the belief value to change accordingly. Therefore, a different belief value should cause the agent to choose a different action, preventing the agent from getting stuck in states.

\subsubsection{Active Learning}

Active learning is a paradigm where the learning agent can query an expert for guidance. It is particularly useful for demonstration learning when the demonstration data set is limited. If the agent is faced with a state that was not present in the demonstration data set, it will not choose the correct action. An incorrect choice may be unsafe. Active learning allows the agent to ask the teacher for an extra demonstration of the correct course of action. 

The approach to update the policy using both demonstration data and the teacher responses requires selecting which option to choose from at any given state. This can be done through a confidence score in \cite{dogged}. If the confidence for a given state-action pair is low, the learning agent will query the teacher for the answer. The learning agent increasingly increases its confidence scores while obtaining a generalized policy. Hence the need for querying the teacher decreases over time. This approach's biggest downside is the requirement of extra investment from the teacher, which in some cases may be unfeasible.
In \cite{ikemoto}, the authors use active learning in human-robot cooperative tasks. For successful cooperation, the robot must be able to adapt its behavior to the human counterpart. Active learning is used after each round of interactions. The expert's feedback is provided by a graphical interface, recorded, and added to a database. This database is then used to update the robot's policy. Results show that the policy of the robot converged more smoothly using this method on standing-up and assisted walking tasks.

% TODO cite d4gers

\subsubsection{Generative Adversarial Imitation Learning (GAIL)}

The authors of \cite{GAIL} introduced a model-free demonstration learning method called Generative Adversarial Imitation Learning (GAIL). These models are inspired by the Generative Adversarial Networks (GAN) and applied to the demonstration learning paradigm. In GAILs, the reward function is learned from the demonstration data and then used in RL for learning the policy.
GANs are composed of two neural networks: a generator and a discriminator. The generator creates new data points and the discriminator has to distinguish generated data points from real data points. The discriminator receives both data points created by the generator and real data points from the data set.
The discriminator must correctly classify the data point as real or generated. 
The accuracy of its prediction will determine how the weights of both networks are adjusted. 
The goal of the generator is to fool the discriminator and is rewarded when successful. Alternatively, the goal of the discriminator is to avoid being fooled and is rewarded for correctly distinguishing real from fake data.

In GAILs, the learned policy $\pi$ performs the role of the generator. The discriminator $D_\phi$ tries to assert whether the state-action pair it received originates from the demonstration data set or from $\pi$. Contrarily, $\pi$ aims to improve its behavior by increasingly approximating it to the one in the demonstration data set, such that its generated trajectories can fool the discriminator into believing they originated from the data set. Therefore, $D_\phi$ is trained as a binary classifier to predict whether the received state-action pair is real or generated. The generator $\pi$ is trained by being rewarded for successfully confusing $D_\phi$, and treating this reward as if it were an external analytically-unknown reward from the environment through RL.

\cite{gail2} create two algorithms for offline GAIL and online GAIL that improve upon state-of-the-art. In \cite{discriminatorweight}, a discriminator is trained to distinguish between two data sets with significant differences in quality. The discriminator is then used as a filter for the policy to avoid learning from sub-optimal data.
In \cite{advskills}, a policy is trained to perform multiple small skills, where each skill is represented by a discriminator, a replay buffer, and a demonstration buffer. Each discriminator learns to differentiate between descriptors built from a pair of consecutive states sampled from either the replay or demonstration buffer. The reward is higher when the policy fools the discriminator into thinking the consecutive states were demonstrated. 

Similarly to GANS, GAILs suffer from severe sample inefficiency. Being sample-inefficient means that the agent can't learn from limited interactions with the environment. This limitation was addressed in later articles such as \cite{SAM}. In \cite{pemirl} and \cite{squirl}, a discriminator is used to distinguish between generated and demonstration state-action pairs to learn multiple similar tasks at once. It is then used to generalize to even more contextually-similar tasks. 

Similar to GAILs, in \cite{zerosum}, a zero-sum game is proposed, where a second player plays the role of an antagonist and perturbs the transition probabilities of the protagonist. 
The antagonist has a perturbation budget which allows for the optimization of the policy for the worst alpha percentile transitions, giving safety guarantees. Recently, \cite{atac} proposes to use adversaries in place of the critic in actor-critic algorithms to improve sample efficiency.

\subsubsection{Embedding Space}

Learning from visual states requires applying a function $f(s)$ that extracts a set of $N$ values, known as features, from the observations: $f(s): S \rightarrow \mathbb{R}^{N}$. Where $N$ is the dimension of the embedding space. With deep learning, these features, and the corresponding embedding space, are typically estimated by applying a set of convolutions and subsampling operations to the input images. In demonstration learning, the states present in the demonstration can be used to explicitly learn an embedding space to extract features with specific characteristics.

Contrastive learning compares different images sharing a common signal to learn representations in a self-supervised fashion.
It has been applied to multiple machine learning fields, most notably image classification \cite{simclr}, where the embedding space is robustly obtained in a self-supervised manner by bringing images from the same class closer in the space. A linear classifier \cite{siamesefeatmatch} is then trained on top of the embedding space for a few epochs to determine the class from the features. 
\cite{clfd} applied this approach to demonstration learning, where demonstrations captured from multiple camera view points are used to estimate a view point invariant embedding space. Then, a view point invariant policy can be obtained, improving the robustness of the policy to changes to the position of the camera.

In \cite{decoupling}, the representation learning part is decoupled from policy estimation, where the embedding space is estimated by contrasting images that appear close to each other in a sequence of frames. In \cite{curl}, the different views are obtained through transformations applied to the original image. Alternatively, in \cite{actionable} the representations are obtained by contrasting the similarity between the sequence of actions required to reach each contrasting state.
In \cite{tcn, atnet}, an embedding space for view-invariant features is estimated from a multi-view data set through triplet learning. Similarly, in \cite{tcc} the authors train an encoder to estimate similar features for concurrent frames of multi-view synchronized videos. However, the criterion here is cycle consistency, where for two views, a data point is cycle-consistent if the nearest neighbor of its nearest neighbor is the point itself. 

Siamese networks \cite{towardssiamese, siamesefeatmatch} have been paired with contrastive learning where each network outputs the features of a different view. In \cite{towardssiamese}, the different views are obtained through data augmentation and applied to motion simulation tasks.
Other forms of obtaining different views from a single image are by using different image channels. In\cite{contrastivecoding}, the images are converted into the Lab color space, where the L and ab components are treated as two views of the image. Then contrastive loss is applied to learn an embedding space. Additionally, they show that increasing the number of views leads to better features.

In \cite{cic} and in \cite{proto} exploration is performed to estimate an embedding space without task-specific returns. The embedding space is then fine-tuned for a specific task using its rewards. The embeddings are encouraged to represent skills by maximizing the mutual information between state transitions and the associated embedding. This encouragement is done through an estimation of the lower bound of this mutual information through the means of a contrastive loss. To explore the agent is trained to maximize accumulated rewards which are proportional to the entropy of state transitions.

In \cite{bissimulation}, the authors use bisimulation to generate an embedding space where functionally identical states from different tasks are mapped to the same embedding. Using this embedding space, the learning agent is trained to adapt to different tasks that are functionally identical to previously learned tasks. 

\subsubsection{Sequence models}
Sequence models learn from a set of transitions instead of a single transition. Because the model has access to the history of past transitions it has more information to make the correct decision. This means that a sequence modeling objective is less prone to deviate from the distribution by selecting out-of-distribution actions and hence less prone to distributional shift.
In sequence models, the objective is to optimize the model over entire trajectories $\pi(\tau)$, and find the best distribution of actions over trajectories.

Transformer networks have revolutionized language processing tasks since their introduction \cite{attisall}. They were first applied to demonstration learning methods in \cite{dectransformer}, where the GPT-3 network processes the sequences of states, actions, and rewards to predict the next action. Later in \cite{onlinedt}, the authors fine-tune the original decision transformer on online data. In \cite{transpretrain}, it is shown that pre-training the decision transformer on a large corpus of language data increases the performance in DL/RL tasks, even though the two have nothing in common at first glance. 
Recently, \cite{hdt} proposed a hierarchical dual-transformer architecture to remove the need for user interaction and improve the performance of sequence models in long horizon tasks. The high-level transformer identifies sub-goal states from demonstrations. The low-level transformer is conditioned on each sub-goal until it completes the task.

\subsection{Multi Agent}

Cooperation between agents is useful for robotic tasks and has been explored in reinforcement learning. However, it has not received the same attention in demonstration learning. 
Most multi-agent research is focused on transferring the skill from one agent to another by having the learner observe the behavior of the teacher. Multi-agent learning increases the complexity of the problem.

The state space must be expanded to include the status of all
agents, as the decisions of an agent, are dependent on the status of the others. The reward function will likely depend on the status of all agents. The reward function can be used in a cooperative setting where the agents aim to maximize the cumulative reward or in a competitive setting where one agent aims to maximize its own reward while minimizing the rewards of others.

In \cite{soccer}, the team of robots collaborates to keep the ball from the other team in a soccer game. Here all the robots share the same policy that is updated after an action from any agent. Hence, the algorithm is the same as single-agent learning.
Alternatively, in \cite{raza} each of the agents learns different roles separately that in the end complement each other.
However, true multi-agent cooperative learning remains an open problem.  

\subsection{Learning Modifications}
In this section, we will discuss modifications that have been employed by demonstration learning algorithms to learning methods as ways to tackle the problems that plague them.

% direct constraints
Constraints are loss terms that are used to restrict the learned model to have specific characteristics. These terms are either distribution constraints or action constraints.
Most commonly, both types of constraints are used to restrict the model to remain inside the distribution of the data set, to reduce the distributional shift and its consequences. Constraint methods can be grouped into direct or indirect. Direct methods estimate the policy of the data set through behavior cloning $\pi_{demo}$ and use it to constrain the learned policy $\pi_\theta$, such that the divergence between the distributions of the two policies is below a threshold $\epsilon$: $|\Delta_{\pi_\theta} - \Delta_{\pi_{demo}}| < \epsilon$. The problem with such constraints is the requirement of the behavior policy. Estimating the behavior policy is difficult due to its reliance on the quality of the demonstration data set. Then, an incorrect behavior policy can cause methods that use it to constrain the learning process to fail. For example, a behavior policy that was learned from sub-optimal or incorrect demonstrations will constrain the policy learning method on such states, causing the policy to be too pessimistic which is undesirable. Furthermore, if more demonstrations become available, direct constraints require re-estimating the behavior policy. In \cite{bcq}, the algorithm estimates the behavior policy using a parametric generative model and constrains the learning policy to make sure it only chooses actions that the behavior policy would choose. Later in \cite{bear}, the authors argue that since constraining the distribution does not take into account the quality of the actions, action constraining is superior. In \cite{brac}, the authors applied a value penalty in the state-value function to improve performance. 

% indirect constraints
Contrarily, indirect methods do not estimate the behavior policy and instead modify the learning objective and use samples from the demonstration data set. The most common way is to minimize the Kullback-Leibler (KL) divergence between the distribution of the learned model and the distribution of the demonstration data set. In \cite{awr,awac}, the algorithms estimate advantage functions to constrain the policy to reduce variance and increase sample efficiency. In \cite{fujimoto}, the authors add a regularizer based on behavior cloning by penalizing the difference between actions from the learned policy and the data set. 

% regularization
Alternatively, instead of imposing constraints, other methods can incentivize the model to have specific behaviors independent of the demonstrated data set. If the regularization term is $\mathbb{T}$, the learning objective changes to incorporate the regularization term: $J'(\pi) = J(\pi) + \mathbb{T}$. Examples of regularization terms include penalizing the weights of the networks \cite{deepq}, and state entropy \cite{proto}, to avoid over-fitting. In \cite{sac}, an entropy regularization term is proposed to control the stochasticity of the policy and promote exploration. Adding entropy, prevents premature convergence, improving robustness and stability. In \cite{cql}, the method learns a lower bound of the true Q-function by adding a regularization term in its estimation.

% uncertainty estimation
Next, we can relax the constraints and regularization based on how much we trust the model. For example, if we estimate the uncertainty of the model, we can reduce safety constraints in low-uncertainty state regions: $J'(\pi) = J(\pi) + \sigma \mathbb{T}$, where $\sigma$ is a function which weights how much to emphasize the regularizing term. A similar weight can be applied for constraints. Entropy estimation methods can be applied as regularization terms such as clustering the state space in \cite{proto} or ensembles of models \cite{agarwal}.

\begin{table}[tb]
\caption{Categorization of Demonstration Learning papers. \label{paperstab}}

\resizebox{\textwidth}{!}{
\begin{tabular}{llrlllllll}
\textbf{Name}             & \textbf{}                                         & \multicolumn{1}{l}{\textbf{Year}} & \textbf{\begin{tabular}[c]{@{}l@{}}Demonstration\\ Technique\end{tabular}} & \textbf{\begin{tabular}[c]{@{}l@{}}Data\\ Representation\end{tabular}} & \textbf{\begin{tabular}[c]{@{}l@{}}Learned\\ Goal\end{tabular}} & \textbf{\begin{tabular}[c]{@{}l@{}}Classification/\\ Regression\end{tabular}} & \textbf{Evaluation Metrics}                                & \textbf{Benchmark}                               & \textbf{Application}   \\ \hline
Abeel et al. 2004         & \cite{abbeel2004}                & 2004                              & Teleoperation                                                              & Raw Data                                                               & IRL                                                             & Regression                                                                    & Acc. Reward                                                & Grid World Car Driving Simulation                & Simulated Tasks        \\
ABPS                      & \cite{androids}                  & 2021                              & N/A                                                                        & N/A                                                                    & Policy Learning                                                 & Classification                                                                & Acc. Reward                                                & Custom Grid World                                & Simulated Tasks        \\
Align-RUDDER              & \cite{alignrudder}               & 2022                              & N/A                                                                        & Raw Data                                                               & Policy Learning + IRL                                           & Regression                                                                    & Succ. Rate                                                 & Minecraft                                        & Simulated Tasks        \\
AOG                       & \cite{wearable}                  & 2017                              & Sensors on Teacher                                                         & Sensor data + Image                                                    & Classification                                                  & Classification                                                                & Succ. Rate                                                 & Water bottle opening                             & Baxter                 \\
APE-V                     & \cite{adaptivepolicies}          & 2022                              & Teleoperation                                                              & Raw Data                                                               & Policy Learning                                                 & Regression                                                                    & Acc. Reward, Succ. Rate                                    & D4RL, Procgen Mazes                              & Simulated Tasks        \\
APID                      & \cite{farahmand}                 & 2013                              & Teleoperation                                                              & Raw Data                                                               & Policy Learning                                                 & Regression                                                                    & Time, Acc. Reward                                          & Path Finding                                     & Simulated, Real Robot  \\
AQuaDem                   & \cite{continuouscontroldiscrete} & 2021                              & Teleoperation                                                              & Raw Data                                                               & Policy Learning                                                 & Classification                                                                & Acc. Reward, Succ. Rate                                    & D4RL                                             & Simulated Tasks        \\
ARC                       & \cite{actionable}                & 2018                              & N/A                                                                        & Image                                                                  & Policy Learning + IRL                                           & Regression                                                                    & Acc. Reward                                                & Navigation, Robot Pushing                        & Simulated Tasks        \\
ATAC                      & \cite{atac}                      & \multicolumn{1}{l}{2022}          & Teleoperation                                                              & Raw Data                                                               & Policy Learning                                                 & Regression                                                                    & Acc. Reward                                                & D4RL                                             & Simulated Tasks        \\
ATC                       & \cite{decoupling}                & 2021                              & Observation                                                                & Image                                                                  & Policy Learning                                                 & Regression                                                                    & Acc. Reward, Train Time                                    & DM control, Atari, DM Lab                        & Simulated Tasks        \\
AT-Net                    & \cite{atnet}                     & 2020                              & Teleoperation                                                              & Image                                                                  & Policy Learning                                                 & Regression                                                                    & Alignment Error, Accuracy, Succ. Rate                      & Manipulation                                     & Simulated Robot        \\
AWAC                      & \cite{awac}                      & \multicolumn{1}{l}{2020}          & Teleoperation                                                              & Raw Data                                                               & Policy Learning                                                 & Regression                                                                    & Acc. Reward + Succ. Rate                                   & Simulated + Real Robot Manipulation              & Sawyer Robot           \\
AWR                       & \cite{awr}                       & \multicolumn{1}{l}{2019}          & Teleoperation                                                              & Raw Data                                                               & Policy Learning                                                 & Regression                                                                    & Acc. Reward                                                & OpenAI Gym, Simulated Robot                      & Simulated Tasks        \\
BCQ                       & \cite{bcq}                       & 2019                              & Teleoperation                                                              & Raw Data                                                               & Policy Learning                                                 & Regression                                                                    & Acc. Reward                                                & OpenAI Gym MuJoCo tasks                          & Simulated Tasks        \\
BEAR                      & \cite{bear}                      & \multicolumn{1}{l}{2019}          & Teleoperation                                                              & Raw Data                                                               & Policy Learning                                                 & Regression                                                                    & Acc. Reward                                                & OpenAI Gym MuJoCo                                & Simulated Tasks        \\
BRAC                      & \cite{brac}                      & \multicolumn{1}{l}{2019}          & Teleoperation                                                              & Raw Data                                                               & Policy Learning                                                 & Regression                                                                    & Acc. Reward                                                & OpenAI Gym MuJoCo                                & Simulated Tasks        \\
BREMEN                    & \cite{bremen}                    & 2020                              & Teleoperation                                                              & Raw Data                                                               & Model-Based Policy Learning                                     & Regression                                                                    & Acc. Reward, KL Divergence                                 & OpenAI Gym MuJoCo                                & Simulated Tasks        \\
CDS                       & \cite{conservativedatasharing}   & 2021                              & Teleoperation                                                              & Raw Data                                                               & Policy Learning                                                 & Regression                                                                    & Acc. Reward, KL Divergence                                 & MuJoCo, Meta-world                               & Simulated Tasks        \\
ChauffeurNet              & \cite{bansal}                    & 2018                              & Teleoperation                                                              & Sensor data                                                            & Policy Learning                                                 & Regression                                                                    & Distance Error                                             & CARLA driving simulator                          & Simulated Tasks        \\
CIC                       & \cite{cic}                       & 2022                              & N/A                                                                        & N/A                                                                    & Policy Learning                                                 & Regression                                                                    & Acc. Reward                                                & Mujoco, Simulated Jaco Robot Manipulation        & Simulated Tasks        \\
CLfD                      & \cite{clfd}                      & 2022                              & Observation                                                                & Image                                                                  & Policy Learning                                                 & Regression                                                                    & Alignment Error, Success Rate, Acc. Reward                 & Simulated Panda Manipulation                     & Simulated Tasks        \\
Coarse-to-Fine IL         & \cite{coarse}                    & 2021                              & Observation                                                                & Image                                                                  & Policy Learning                                                 & Regression                                                                    & Error, Succ. Rate                                          & Target Reaching                                  & Sawyer                 \\
Codevilla et al., 2018    & \cite{codevilla}                 & 2018                              & Teleoperation                                                              & Sensor data                                                            & Policy Learning                                                 & Regression                                                                    & Succ. Rate, \#Missed turns, \#Interventions, \#Infractions & Real Truck Driving                               & Real Truck             \\
Confidence-Based LfD      & \cite{veloso}                    & 2007                              & Teleoperation                                                              & Raw Data                                                               & GMM                                                             & Classification                                                                & Accuracy, Collision Rate                                   & Custom Simulation                                & Simulated Tasks        \\
Context-Aware Translation & \cite{gupta}                     & 2018                              & Observation                                                                & Image                                                                  & Policy Learning,IRL                                             & Regression                                                                    & Error, Succ. Rate                                          & MuJoCo Manipulation                              & Simulated Tasks        \\
COPO                      & \cite{copo}                      & 2022                              & Teleoperation                                                              & Raw Data                                                               & Policy Learning                                                 & Regression                                                                    & Acc. Reward, Cost, \#Violations                            & Walk-Around-Grid, Bipedal Walker                 & Simulated Tasks        \\
COMBO                     & \cite{combo}                     & 2021                              & Teleoperation                                                              & Raw Data                                                               & Model-Based Policy Learning                                     & Regression                                                                    & Acc. Reward, Model Error                                   & D4RL                                             & Simulated Tasks        \\
CORRO                     & \cite{corro}                     & 2022                              & Teleoperation                                                              & Raw Data                                                               & Policy Learning                                                 & Regression                                                                    & Acc. Reward                                                & Mujoco                                           & Simulated Tasks        \\
CQL                       & \cite{cql}                       & 2020                              & Teleoperation                                                              & Raw Data                                                               & Policy Learning                                                 & Regression                                                                    & Acc. Reward, Value Error                                   & D4RL                                             & Simulated Tasks        \\
Cross-Context IL          & \cite{crosscontext}              & 2020                              & Observation                                                                & Image                                                                  & Policy Learning                                                 & Regression                                                                    & Distance, Succ. Rate                                       & Manipulation                                     & UR5                    \\
CSI                       & \cite{csi}                       & 2013                              & Teleoperation                                                              & Raw Data                                                               & IRL                                                             & Both                                                                          & Acc. Reward                                                & Mountain Car, Driving Simulator                  & Simulated Tasks        \\
CVaR                      & \cite{zerosum}                   & 2022                              & N/A                                                                        & N/A                                                                    & Policy Learning                                                 & Regression                                                                    & Acc. Reward                                                & Custom Grid World                                & Simulated Tasks        \\
CVPO                      & \cite{constraint}                & 2022                              & N/A                                                                        & N/A                                                                    & Policy Learning                                                 & Regression                                                                    & Cost, Acc. Reward                                          & Custom Simulation Tasks                          & Simulated Tasks        \\
DAgger                    & \cite{mario}                     & 2011                              & Teleoperation                                                              & Raw Data                                                               & Policy Learning                                                 & Classification                                                                & \#Falls, Distance Traveled                                 & Super Tux Kart, Super Mario Bros.                & Simulated Tasks        \\
Dalal et al., 2018        & \cite{safecont}                  & 2018                              & Teleoperation                                                              & Raw Data                                                               & Policy Learning                                                 & regression                                                                    & Acc. Reward, \#Violations                                  & Custom MuJoCo tasks                              & Simulated Tasks        \\
DIS                       & \cite{shaping}                   & 2016                              & Teleoperation                                                              & Raw Data                                                               & Policy Learning, IRL                                            & Regression                                                                    & Acc. Reward                                                & Maze, Mario AI                                   & Simulated Tasks        \\
DQfD                      & \cite{deepq}                      & \multicolumn{1}{l}{2018}          & Teleoperation                                                              & Image                                                                  & Policy Learning                                                 & Classification                                                                & Acc. Reward                                                & ALE                                              & Simulated Tasks        \\
DQN                       & \cite{dqn}                       & 2015                              & N/A                                                                        & N/A                                                                    & Policy Learning                                                 & Classification                                                                & Acc. Reward                                                & Atari                                            & Simulated Tasks        \\
Dogged Learning           & \cite{dogged}                    & 2007                              & Teleoperation                                                              & Raw Data, Image                                                        & Other  Predictive Learning                                      & Either or                                                                     & Student visual inspection, Error                           & Ball seeking, head mirroring tail                & Sony Aibo robot        \\
DoubIL/ResiduIL           & \cite{causal}                    & 2022                              & Teleoperation                                                              & Raw Data                                                               & Policy Learning                                                 & Regression                                                                    & Loss                                                       & OpenAI Gym                                       & Simulated Tasks        \\
DT                        & \cite{dectransformer}            & 2021                              & Teleoperation                                                              & Raw Data                                                               & Sequence Model                                                  & Regression                                                                    & Acc. Reward                                                & Atari, OpenAI Gym                                & Simulated Tasks        \\
DWBC                      & \cite{discriminatorweight}       & 2022                              & Teleoperation                                                              & Raw Data                                                               & Policy Learning                                                 & Regression                                                                    & Acc. Reward, Discriminator Accuracy                        & D4RL                                             & Simulated Tasks        \\
EnsembleDAgger            & \cite{ensembledagger}            & 2019                              & N/A                                                                        & N/A                                                                    & Policy Learning                                                 & Regression                                                                    & Acc. Reward                                                & OpenAI Gym                                       & Simulated Tasks        \\
ExORL                     & \cite{yarats}                    & 2022                              & Teleoperation                                                              & Raw Data                                                               & Policy Learning                                                 & Regression                                                                    & Acc. Reward                                                & DeepMind control suite                           & Simulated Tasks        \\
GAIL                      & \cite{GAIL}                      & 2016                              & Teleoperation                                                              & Raw Data                                                               & GAIL                                                            & Regression                                                                    & Acc. Reward                                                & MuJoCo                                           & Simulated Tasks        \\
GCB                       & \cite{bissimulation}             & 2022                              & Observation                                                                & Image                                                                  & Policy Learning                                                 & Regression                                                                    & Succ. Rate                                                 & Pybullet                                         & Simulated Tasks        \\
Gradient-Based IRL        & \cite{fb}                        & 2021                              & Observation                                                                & Image                                                                  & Model Learning, IRL                                             & Regression                                                                    & Train Time, Distance                                       & Teaching                                         & KUKA                   \\
GTI                       & \cite{gti}                       & 2021                              & Teleoperation                                                              & Raw Data                                                               & Policy Learning                                                 & Regression                                                                    & Succ. Rate, Generalization                                 & Manipulation                                     & Panda                  \\
Guo et al., 2022          & \cite{provablyeff}               & 2022                              & Observation                                                                & Image                                                                  & Policy Learning                                                 & Regression                                                                    & N/A                                                        & N/A                                              & N/A                    \\
HAMMER                    & \cite{demiris}                   & 2005                              & Observation                                                                & Image                                                                  & Bayesian Belief                                                 & Regression                                                                    & N/A                                                        & N/A                                              & N/A                    \\
Hayes et al., 2014        & \cite{legos}                     & 2014                              & Observation                                                                & Image                                                                  & Active learning                                                 & Regression                                                                    & Execution Paths                                            & Lego Montage                                     & Manipulator            \\
HDT                       & \cite{hdt}                       & \multicolumn{1}{l}{2022}          & Teleoperation                                                              & Raw Data                                                               & Sequence Model                                                  & Regression                                                                    & Acc. Reward                                                & UR3 Reaching, D4RL                               & UR3                    \\
IRIS                      & \cite{iris}                      & 2020                              & Teleoperation                                                              & Raw Data                                                               & Policy Learning                                                 & Regression                                                                    & Succ. Rate, Acc. Reward, Traj. Length                      & Graph Reach, RoboTurk, Robosuite                 & Simulated Tasks        \\
Ijspeert et al., 2002     & \cite{speert}                    & 2002                              & Sensors on Teacher                                                         & Sensor Data                                                            & LWR                                                             & Regression                                                                    & Error                                                      & Tennis Swings                                    & Humanoid Robot         \\
ILEED                     & \cite{expertise}                 & 2022                              & Teleoperation                                                              & Raw Data                                                               & Policy Learning                                                 & Regression                                                                    & Acc. Reward                                                & Simulated Minigrid tasks                         & Simulated Tasks        \\
LazyDAgger                & \cite{lazydagger}                & 2021                              & N/A                                                                        & N/A                                                                    & Policy Learning                                                 & Regression                                                                    & Acc. Reward, \#Interventions                               & MuJoCo                                           & Simulated Tasks        \\
Levine et al., 2016       & \cite{levine}                    & 2016                              & N/A                                                                        & Image                                                                  & Policy Learning                                                 & Regression                                                                    & Distance, Succ. Rate, Error                                & Manipulation                                     & PR2                    \\
Levine et al. 2018        & \cite{levine2018}                & 2018                              & Observation                                                                & Image                                                                  & Policy Learning                                                 & Regression                                                                    & Failure Rate                                               & Manipulation Task                                & KUKA arms              \\
LDM                       & \cite{ldm}                       & 2022                              & Teleoperation                                                              & Raw Data                                                               & Policy Learning                                                 & Regression                                                                    & Acc. Reward, Succ. Rate                                    & OpenAI Gym MuJoCo, SimGlucose                    & Simulated Tasks        \\
LTSD                      & \cite{ltsd}                      & 2019                              & Teleoperation                                                              & Raw Data                                                               & Policy Learning, IRL                                            & Regression                                                                    & Acc. Reward                                                & BiMGame, AntTarget, AntMaze                      & Simulated Tasks        \\
LfMD                      & \cite{knot}                      & 2015                              & Teleoperation                                                              & Point Cloud                                                            & Other                                                           & N/A                                                                           & Succ. Rate                                                 & Pick Place, Towel Folding                        & PR2 robot              \\
Maeda et al., 2017        & \cite{kinesthetic}               & 2017                              & Kinesthetic                                                                & Sensor Data                                                            & Policy Learning                                                 & Regression                                                                    & Distance Error                                             & Manipulation Tasks                               & KUKA                   \\
MAGIC                     & \cite{brunskill}                 & 2016                              & N/A                                                                        & N/A                                                                    & Policy Learning (OPE)                                           & Regression                                                                    & Reg. Error                                                 & ModelFail,ModelWin,Maze, Mountain Car, Cart Pole & Simulated Tasks        \\
Max. Ent. IRL             & \cite{ziebart}                   & 2008                              & Sensors                                                                    & GPS Data                                                               & IRL                                                             & N/A                                                                           & Matching                                                   & Path Following                                   & Mobile Robot           \\
MBPO                      & \cite{janner}                    & 2019                              & N/A                                                                        & N/A                                                                    & Policy Learning (OPE)                                           & Regression                                                                    & Acc. Reward, Reg. Error                                    & MuJoCo                                           & Simulated Tasks        \\
MERLION                   & \cite{pseudometric}              & 2021                              & Teleoperation                                                              & Raw Data                                                               & Policy Learning                                                 & Classification                                                                & Acc. Reward                                                & Maze, Dialogue, Recommendation                   & Simulated Tasks        \\
MIR                       & \cite{manipulatorindependent}    & 2021                              & Observation                                                                & Image                                                                  & Policy Learning                                                 & Regression                                                                    & Succ. Rate                                                 & MuJoCo, Manipulation                             & Sawyer                 \\
MOPO                      & \cite{mopo}                      & 2020                              & Teleoperation                                                              & Raw Data                                                               & Model-Based Policy Learning                                     & Regression                                                                    & Acc. Reward                                                & D4RL                                             & Simulated Tasks        \\
MOReL                     & \cite{kidambi}                   & 2020                              & Teleoperation                                                              & Raw Data                                                               & Model-Based Policy Learning                                     & Regression                                                                    & Acc. Reward                                                & OpenAI Gym MuJoCo                                & Simulated Tasks        \\
Motion2Vec                & \cite{seqrobotil}                & 2021                              & Kinesthetic+Observation                                                    & Robot Data+Image                                                       & HMM                                                             & Regression                                                                    & Loss, Segmentation Accuracy, Noise                         & Pick-and-Place, Suturing                         & Baxter                 \\
MRDR                      & \cite{farajtabar}                & 2018                              & N/A                                                                        & N/A                                                                    & Policy Learning (OPE)                                           & Regression                                                                    & Reg. Error                                                 & ModelFail,ModelWin,Maze, Mountain Car, Cart Pole & Simulated Tasks        \\
Mülling et al., 2013      & \cite{tennis}                    & 2013                              & Kinesthetic                                                                & Raw Data                                                               & LWR                                                             & Regression                                                                    & Cost, Succ. Rate, Acc. Reward                              & Tennis Swings                                    & Robot Arm              \\
Multi-AMP                 & \cite{advskills}                 & 2022                              & Observation                                                                & Image                                                                  & Policy Learning                                                 & Regression                                                                    & Stand Duration                                             & 4 Legged Robot Movements                         & Real Quadruped Robot   \\
MVP                       & \cite{maskedpre}                 & 2022                              & Observation                                                                & Image                                                                  & Policy Learning                                                 & Regression                                                                    & Succ. Rate                                                 & PixMC                                            & Simulated KUKA, Franka \\
ODT                       & \cite{onlinedt}                  & 2022                              & N/A                                                                        & Raw Data                                                               & Policy Learning                                                 & Regression                                                                    & Acc. Reward                                                & D4RL                                             & Simulated Tasks        \\
Pan et al., 2017          & \cite{pan}                       & 2017                              & Teleoperation                                                              & Sensor data + Image                                                    & Policy Learning                                                 & Regression                                                                    & Speed, Succ. rate                                          & AutoRally                                        & AutoRally              \\
PC-GMM                    & \cite{learningsoft}              & 2020                              & Teleoperation                                                              & Raw Data                                                               & GMM, GMR, RL                                                    & Regression                                                                    & Succ. Rate, Error                                          & Peg-in-Hole                                      & UR5                    \\
PEMIRL                    & \cite{pemirl}                    & 2019                              & Teleoperation                                                              & Raw Data                                                               & IRL                                                             & Regression                                                                    & Acc. Reward                                                & MuJoCo                                           & Simulated Tasks        \\
PI2-ES-Cov                & \cite{fewdemos}                  & 2020                              & Kinesthetic                                                                & features from point cloud                                              & Policy Learning                                                 & Regression                                                                    & Succ. Rate                                                 & Simulated + Real Robot Manipulation              & Franka Emika Panda     \\
Pinto et al., 2016        & \cite{pinto2016}                 & 2016                              & Observation                                                                & Image                                                                  & Policy Learning                                                 & Regression                                                                    & Accuracy                                                   & Manipulation Task                                & Baxter                 \\
PTS                       & \cite{brys}                      & 2015                              & N/A                                                                        & N/A                                                                    & IRL                                                             & Regression                                                                    & Score                                                      & Mounting Car 3d, Cart Pole, Mario                & Simulated Tasks        \\
Recovery RL               & \cite{recovery}                  & 2021                              & Teleoperation                                                              & Raw Data                                                               & Policy Learning                                                 & Regression                                                                    & Succ. Rate, \#Violations                                   & Simulated task, Real Robot                       & Simulated, Real Robot  \\
REPaIR                    & \cite{repair}                    & 2020                              & Teleoperation                                                              & Raw Data                                                               & Policy Learning                                                 & Regression                                                                    & Acc. Reward, AUC, Accuracy                                 & Robot Reaching                                   & Kinova Jaco            \\
Rhinehart et al., 2018    & \cite{rhinehart}                 & 2018                              & Teleoperation                                                              & Raw Data                                                               & Model-Based Policy Learning                                     & Regression                                                                    & Succ. Rate, Classification Metrics                         & CARLA driving simulator                          & Simulated Tasks        \\
RLfD                      & \cite{shaping2}                  & 2015                              & Teleoperation                                                              & Raw Data                                                               & Policy Learning, IRL                                            & Regression                                                                    & Acc. Reward                                                & Mario AI, Cart Pole                              & Simulated Tasks        \\
RIS                       & \cite{ris}                       & 2021                              & N/A                                                                        & N/A                                                                    & Policy Learning                                                 & Regression                                                                    & Acc. Reward, Distance, Succ. Rate                          & Navigation, Manipulation                         & Simulated Tasks        \\
RPL                       & \cite{relay}                     & 2019                              & Teleoperation                                                              & Raw Data                                                               & Policy Learning                                                 & Regression                                                                    & Succ. Rate                                                 & MuJoCo                                           & Simulated Tasks        \\
RUDDER                    & \cite{rudder}                    & 2019                              & N/A                                                                        & Raw Data                                                               & Policy Learning + IRL                                           & Regression                                                                    & Acc. Reward                                                & Atari                                            & Simulated Tasks        \\
Ruppel et al., 2020       & \cite{customglove}               & 2020                              & Sensors on Teacher                                                         & Sensor Data                                                            & Sequence Model                                                  & Regression                                                                    & Succ. Rate, Error                                          & Manipulation                                     & UR10e, UR10, KUKA      \\
S4RL                      & \cite{s4rl}                      & 2022                              & Teleoperation                                                              & Image                                                                  & Policy Learning                                                 & regression                                                                    & Acc. Reward                                                & D4RL, MetaWorld, RoboSuite                       & Simulated Tasks        \\
SafeDAgger                & \cite{safedagger}                & 2017                              & Teleoperation                                                              & Raw Data                                                               & Policy Learning                                                 & Regression                                                                    & Score, Safety, loss                                        & TORCS Driving Car                                & Simulated Tasks        \\
SAILR                     & \cite{intervention}              & 2021                              & N/A                                                                        & N/A                                                                    & Policy Learning                                                 & Regression                                                                    & Acc. Reward, Constraints                                   & Point Robot, OpenAI Gym MuJoCo                   & Simulated Tasks        \\
SAM                       & \cite{SAM}                       & 2019                              & Teleoperation                                                              & Raw Data                                                               & GAIL                                                            & Regression                                                                    & Acc. Reward                                                & MuJoCo                                           & Simulated Tasks        \\
SCIRL                     & \cite{scirl}                     & 2012                              & Teleoperation                                                              & Raw Data                                                               & IRL                                                             & Classification                                                                & Acc. Reward                                                & Highway                                          & Simulated Tasks        \\
Sepsis Treatment          & \cite{raghu}                     & 2017                              & N/A                                                                        & N/A                                                                    & Policy Learning                                                 & Classification                                                                & Mortality                                                  & MIMIC-3.                                         & N/A                    \\
Silver et al., 2010       & \cite{terrain}                   & 2010                              & Kinesthetic                                                                & Raw Data                                                               & LWR                                                             & Regression                                                                    & Cost, Loss, Distance, Speed                                & Navigation                                       & Mobile Robot           \\
SMILe                     & \cite{smile}                     & 2010                              & Teleoperation                                                              & Raw Data                                                               & Policy Learning                                                 & Regression                                                                    & \#Falls, Distance                                          & Mario Bros                                       & Simulated Tasks        \\
SPiRL                     & \cite{accelarating}              & 2020                              & Teleoperation                                                              & Raw Data                                                               & Policy Learning                                                 & Regression                                                                    & Succ. Rate                                                 & D4RL                                             & Simulated Tasks        \\
SQUIRL                    & \cite{squirl}                    & 2020                              & Observation                                                                & Image                                                                  & Policy Learning, IRL                                            & Regression                                                                    & Succ. Rate                                                 & Pick-Carry-Drop, Pick-Place                      & UR5                    \\
SRL-RNN                   & \cite{drug}                      & 2018                              & N/A                                                                        & N/A                                                                    & Sequence Model                                                  & Both                                                                          & Estimated Mortality                                        & MIMIC-3.                                         & Simulated Tasks        \\
SWIRL                     & \cite{swirl}                     & 2019                              & Teleoperation                                                              & Raw Data                                                               & Policy Learning                                                 & Regression                                                                    & Acc. Reward                                                & RCCar, Acrobot                                   & Simulated Tasks        \\
SWITCH                    & \cite{wang2017}                  & 2017                              & N/A                                                                        & N/A                                                                    & Policy Learning (OPE)                                           & Regression                                                                    & Reg. Error                                                 & UCI data sets                                    & Simulated Tasks        \\
TCC                       & \cite{tcc}                       & 2019                              & Observation                                                                & Image                                                                  & Representation Learning                                         & N/A                                                                           & Classification Accuracy, Kendall's Tau                     & Pouring and Penn action data sets                & N/A                    \\
TCN                       & \cite{tcn}                       & 2018                              & Observation                                                                & Image                                                                  & Policy Learning                                                 & Regression                                                                    & Alignment Error, Classification Error, Acc. Reward         & Pouring                                          & KUKA                   \\
TD3+BC                    & \cite{fujimoto}                  & 2021                              & Teleoperation                                                              & Raw Data                                                               & Policy Learning                                                 & Regression                                                                    & Acc. Reward                                                & D4RL                                             & Simulated Tasks        \\
TPIL                      & \cite{disentagle}                & 2021                              & Observation                                                                & Image                                                                  & Policy Learning                                                 & Regression                                                                    & Alignment Error, Loss                                      & PyBullet, Minecraft                              & Panda                  \\
UCT                       & \cite{guo}                       & 2014                              & Teleoperation                                                              & Image                                                                  & Regressor/Classifier                                            & Both                                                                          & Score                                                      & ALE                                              & Simulated Tasks        \\
UDS                       & \cite{leverage}                  & 2022                              & Teleoperation                                                              & Raw Data                                                               & Policy Learning                                                 & Regression                                                                    & Acc. Reward                                                & D4RL                                             & Simulated Tasks        \\
visual MPC                & \cite{ebert}                     & 2018                              & Observation                                                                & Image                                                                  & Model-Based policy learning                                     & Regression                                                                    & Reg. Error, Succ. Rate,                                    & Manipulation Task                                & Robot Arm              \\
Weak Label LfD            & \cite{alien}                     & 2020                              & Teleoperation                                                              & Sensor Data + Image                                                    & Classifier                                                      & Classification                                                                & Effort                                                     & Manipulation                                     & PR2                    \\
Yang et al., 2022         & \cite{regularizing}              & 2022                              & Teleoperation                                                              & Raw Data                                                               & Policy Learning + GAN                                           & Regression                                                                    & Acc. Reward                                                & D4RL                                             & Simulated Tasks        \\ \hline
\end{tabular}
}
\end{table}

\section{Evaluation}

\begin{table}[!t]
\caption{Distinction between evaluation methods.}
\label{evaltab}
\centering
\begin{tabular}{ccc}
                   & \textbf{Quantitative} & \textbf{Qualitative} \\ \hline
\textbf{Goal}      & Performance           & Believability        \\
\textbf{Judgement} & Objective             & Subjective           \\
\textbf{Metric}    & Distance to goal      & Subjective Analysis 
\end{tabular}

\end{table}

Like other machine learning paradigms, demonstration learning methods need to be evaluated using a set of metrics. The evaluation metrics are split into quantitative and qualitative, and the distinctions between the two are summarized in Table \ref{evaltab}. Demonstration learning inherits the metrics from reinforcement and supervised learning. The most common performance metrics are the success rate, accumulated reward, and classification or regression error on a demonstration test set. However, some applications prioritize human-like behavior. Additionally, demonstration learning suffers from the problem of specifying the values of hyper-parameters.

It is common that experiments are conducted on specific robots or simulators designed for the method. Due to the limited number of benchmarks, it can be difficult to perform evaluations.
Demonstration learning can use reinforcement learning benchmark simulation environments. Some include demonstration data sets. However, even if this is not the case, one can train a policy through reinforcement learning to learn the simulation task, and then use the policy to generate demonstration data sets. Furthermore, bench-marking in the real world is still complicated. Because of the required hardware, different backgrounds, and safety concerns, usually a custom task environment and corresponding demonstration data sets need to be created for each individual method. However, some real-world demonstration data sets exist. This section elaborates on the topics of evaluation in demonstration learning.

\subsection{Quantitative}

The quantitative evaluation metrics are specific to tasks where the performance of a policy can be measured directly. 
The quantitative evaluation approaches are split into two categories: online and off-policy evaluation. Online evaluation is the most common of the two. After training the policy, a set of $N$ online rollouts ${\tau_1, ..., \tau_N}$ are performed on the environment and a metric is used on such rollouts. A rollout is the sequence of transitions generated by selecting the action from the current policy and obtaining the next state and reward from the transition and reward functions, respectively: $\tau_i = {(s_0,a_0,r_0), ..., (s_H,a_H,r_H)}$, where $H$ is the length of the trajectory. The most common measurement of online performance is the average success rate of the policy at performing the task $J(\pi) = \frac{\sum_{i=0}^{N} \mathbb{G}(\tau_i)} {N}$, where $\mathbb{G}$ is 1 if the trajectory completed the task and 0 otherwise. For example in \cite{kober}, the success depends on if the ball falls inside the cup or outside.

If a reward function is available, the performance of the model can be measured by the accumulated rewards of a rollout \cite{deepq}: $J(\pi) = \sum_{t=0}^{H} R(s_t,a_t,s_{t+1})$. Similarly, we can measure the average or the maximum accumulated rewards over multiple rollouts. A similar metric, mostly used in video games, is obtaining the score directly from the environment and using it as a metric. In \cite{mario}, the performance can be evaluated by the distance traveled. 

Alternatively, if the goal is to imitate the teacher as closely as possible, the measurement could be the distance between the agent and the teacher's actions \cite{schaal2003}, through a classification or regression error, for discrete or continuous action spaces, respectively: $J(\pi) = \sum_{(s_i,a_i)\in D_{demo}} ||\pi(s_i) - a_i||$.

Time can be used as an evaluation metric. For example, the time of execution of the task or the time of convergence during learning, such as the number of training steps. In \cite{torrey} they simulated RoboCup soccer agents where the goal of the task is to keep the ball from the enemy team. Therefore, the duration that the ball is kept from the enemy team can be used as a performance metric. 

Lastly, another way of evaluating the policy's performance online can be by measuring the safety of the method. Safety metrics include the length of the episode (the number of transitions), or defining a set of safety constraints and counting the number of times the agent violated the constraints. This can be done, for example, by associating certain states with violation occurrences or by defining the violations separately from the state space and after the agent executes an action checking if any violation occurred. In \cite{daap}, an advisor agent aims to prevent the main agent from violating constraints that could cause damage to itself while learning.
The problem with online evaluation is that it relies on interactions with the environment which can be dangerous. This fact prevents evaluating a policy while learning it because it might be too dangerous to deploy. 

Demonstration learning inherits problems from machine learning, specifically, determining the ideal values of the hyper-parameters. Determining the ideal values before training or without completing the training saves time and computing resources due to having to run the experiment multiple times varying the set of values. Additionally, the values may avoid dangerous interactions after deployment. 

Off-policy evaluation (OPE) is the process of evaluating a policy using past experience. This past experience can be the demonstration data set, a memory of online interactions, or a combination of both. 
In \cite{paine}, a review of different OPE methods for the selection of hyper-parameter values. Most demonstration learning methods do not use OPE and evaluate performance on a set of hyper-parameters. The decision behind the choice of parameters, in this case, is often from the state-of-the-art, similar past methods or random. Others train the model for multiple sets of hyper-parameter values.

OPE methods depend on a data set of past interactions, $D_{demo}$, and an optimization objective, $J(\pi)$. Some OPE methods use the transition function $P(s_{t+1}\mid s_t,a_t)$ and the reward function $R(s_t,a_t,s_{t+1})$ to evaluate the policy. If these are not available for the current problem, model-based methods estimate dynamics model $\psi(s_t,a_t) \sim P(s_{t+1}\mid s_t,a_t)$ and a reward function through IRL $\hat{R}(s_t,a_t,s_{t+1})$. Using the transition function, reward function, and the actions selected using the current policy $\pi$, we can calculate the expected return: $J(\pi) = \mathbb{E}_{a_t \sim \pi(s_t), s_{t+1} \sim P(s_t,a_t)}[\sum_{t=0}^{H}\gamma^{t} R(s_t,a_t,s_{t+1})]$.

Alternatively, instead of using a transition and reward function, we can estimate a state-action value function $Q(s,a)$ by minimizing the Bellman error using the data set $D_{demo}$ and current policy $\pi$. The OPE objective is the expected accumulated rewards by the state-value function: $J(\pi) = \mathbb{E}_{(s,a) \sim D_{demo}}[Q(s,a)]$.

An expansion from this evaluation method is to weight the importance of each reward differently, using importance sampling \cite{gendice}.
The weights are obtained by first estimating a policy from the demonstration data set through behavior cloning $pi_{BC}$. A form weighting is the product of importance: $w = \frac{\prod_{t=0}^{H} \pi(a_t\mid s_t)}{\prod_{t=0}^{H}\pi_{BC}(a_t\mid s_t)}$. The weights can be used to regulate the previous objective as such: $J(\pi) = \mathbb{E}_{a_t \sim \pi(s_t), s_{t+1} \sim P(s_t,a_t)}[w \sum_{t=0}^{H}\gamma^{t} R(s_t,a_t,s_{t+1})]$.

\subsection{Qualitative}
Qualitative metrics are specific for tasks where the behavior of the agent is more important than what it achieves. As aforementioned, some applications require the agent to simulate how a human would behave. However, it is not easy to objectively quantify the believability of the model. One way of evaluating it is to use multiple judges, each providing their own analysis, and scoring the agent from them.

The authors in \cite{ortega} tested different methods for generating controllers that best-represented human behavior. The methods were evaluated on "Super Mario Bros" game but could be applied to other tasks. The evaluation was done qualitatively. Users are presented with pairs of game-play, one performed by a trained controller and another by a human, and must identify which one corresponds to human game-play, for each pair.

\section{Benchmarks}
\label{benchmarks}

\begin{table}[!]

\caption{Summary of the available benchmarks for demonstration learning methods.}
\label{benchmarktab}

\resizebox{\textwidth}{!}{
\begin{tabular}{llccccc}
\multicolumn{1}{c}{\textbf{Benchmark}} & \multicolumn{1}{c}{\textbf{Data Set}} & \textbf{Action Space} & \textbf{State Space} & \textbf{Dynamics} & \textbf{Observability} & \textbf{Type}           \\ \hline
AI Habitat                             & Included                              & Discrete              & Continuous           & Deterministic     & Full                   & Simulation              \\
Adroit                                 & D4RL                                  & Continuous            & Continuous           & Deterministic     & Full                   & Simulation              \\
ALE                                    & No                                    & Discrete              & Visual/Continuous    & Stochastic        & Full                   & Simulation              \\
Atari                                  & RL Unplugged                          & Discrete              & Visual/Continuous    & Stochastic        & Full                   & Simulation              \\
BSuite                                 & No                                    & Discrete              & Continuous           & Deterministic     & Full                   & Simulation              \\
DM Control                             & RL Unplugged                          & Continuous            & Continuous           & Both              & Full                   & Simulation              \\
DM Lab                                 & No                                    & Continuous            & Visual/Continuous    & Deterministic     & Partial                & Simulation              \\
DM Locomotion                          & RL Unplugged                          & Continuous            & Visual               & Deterministic     & Both                   & Simulation              \\
Google Research Footbal                & No                                    & Discrete              & Continuous           & Deterministic     & Full                   & Simulation              \\
Gym-MuJoCo                             & D4RL                                  & Continuous            & Continuous           & Deterministic     & Full                   & Simulation              \\
Gym-Retro                              & No                                    & Discrete              & Visual/Continuous    & Stochastic        & Full                   & Simulation              \\
Meta-World                             & No                                    & Continuous            & Continuous           & Stochastic        & Full                   & Simulation              \\
Mine-RL                                & Included                              & Both                  & Visual               & Deterministic     & Partial                & Simulation              \\
Real-World Reinforcement Learning      & No                                    & Continuous            & Continuous           & Deterministic     & Full                   & Simulation              \\
RoboTurk                               & Included                              & Continuous            & Continuous           & Stochastic        & Both                   & Real Robot + Simulation
\end{tabular}
}

\end{table}

Often the demonstration data set is obtained from rollouts of a policy previously trained using an online RL algorithm. Alternatively, a human demonstrator can interact with the environment and generate demonstrations.
Hence, online RL benchmarks can be employed for demonstration learning in these cases. 
We summarize the available benchmarks in Table \ref{benchmarktab}.

AI Habitat is a simulation platform for the research and development of embodied agents in an efficient 3D simulator, before transferring the learned skills to reality.
Another benchmark is BSuite \cite{bsuite}, which provides a set of experiments that evaluate the capabilities of a learning method. The library automates the evaluation of an agent in 9 varied environments. 
DeepMind's control suite \cite{dmcontrol} is a benchmark providing a set of Reinforcement Learning environments, built on top of the MuJoCo simulator.
The tasks include controlling a variety of agents: Acrobot, Ball-in-cup, Cart-pole, Cheetah, Finger, Fish, Hopper, Humanoid, Manipulator, Pendulum, Point-mass, Reacher, Swimmer, and Walker.
The state spaces are non-visual.
DeepMind Lab \cite{dmlab} is a 3D learning environment built on the Quake III Arena game. The agents should learn from visual observations, challenging 3D navigation, and puzzle-solving tasks. The most widely used visual benchmark is OpenAI's Gym benchmark. It includes teaching an agent to walk, such as a Humanoid agent, similar to DeepMind's control suite, using MuJoCo. It additionally provides environments for classic Atari games. Gym Retro expands from OpenAI Gym, providing environments for over 1000 classic games.
Google Research Football \cite{googlefootball} is a novel RL environment providing a physics-based 3D football simulation. The agents can control either a single player or the entire team. Hence it can be used for multi-agent and multi-task learning methods.

Meta-World \cite{meta-world} is a benchmark for meta-reinforcement learning and multi-task learning consisting of 50 distinct robotic manipulation tasks. Meta-learning algorithms can acquire new skills more quickly, by leveraging prior experience to learn how to learn. The
Real-World Reinforcement Learning (RWRL) Suite \cite{rwrls} identifies nine challenges that prevent Reinforcement Learning (RL) agents from being applied to the real world. It also describes a framework and a set of environments to evaluate the method's potential applicability to the real world. 

In practical applications, such as the real world, we do not have access to a policy. In such scenarios, the data might come from non-Markovian agents such as humans. Hence, the data sets generated by the online RL policies are not representative of practical applications. There is a continuous growth of available benchmarks in the field, with varying properties. The demonstration data sets should take into account the properties explained in 
Section \ref{data set}. The environment should also take into account certain characteristics. Continuous action and state spaces are often more challenging than discrete ones since it is unfeasible to visit every state and perform every action. Hence, an agent that learns in such domains is obligated to generalize beyond the visited data. Furthermore, visual observations are considered
just more challenging than non-visual observations. A further complication that is common in the real world is the occlusion of the state representation. In such cases, the states are partially observable, and the problem becomes a POMDP. This problem is much harder and closer to the real world where it is difficult to capture all the information that defines a state. Lastly, stochastic transition functions are more common in the real world. Hence, a stochastic environment is more desirable than a deterministic one for evaluating methods. 

The D4RL \cite{d4rl} benchmark provides demonstration data sets for OpenAI Gym's MuJoCo tasks. These data sets come from human demonstrations and vary in quality for example random, medium, and expert.
The RL Unplugged \cite{rlunplugged} benchmark provides demonstration data set for four different suites: Arcade Learning Environment (ALE) \cite{ale}, DeepMind Control Suite \cite{dmcontrol}, DeepMind Locomotion \cite{dmlocomotion}, and
Real-World Reinforcement Learning Suite \cite{rwrls}. The disadvantage is that the data sets of RL Unplugged are obtained from online RL policies. This means they do not include the desirable non-Markovian properties which arise from human demonstrations.

The Deep Off-Policy Evaluation \cite{fu} (DOPE) benchmark provides a standardized framework for comparing OPE algorithms. 
The benchmark provides six OPE methods and is divided into two suites: DOPE RL Unplugged and DOPE D4RL. DOPE allows the selection of different learning objectives, such as ensuring the estimated
value of a policy is as close as possible to the true value, selecting the best possible policy from a set of policies, and hyper-parameter tuning with early stopping.

MineRL \cite{minerl} is a benchmark built on top of the game Minecraft. The benchmark includes many tasks such as finding a cave, creating a pen, making a waterfall, building a house, and finding a diamond. Because the tasks are a sequence of steps, the benchmark requires some of the hardest problems for learning methods, such as sparse rewards, and long horizons.
The benchmark provides a large imitation learning data set with over 60 million frames of recorded human player data.

For real robots, RoboTurk \cite{roboturk} provides the largest collection of demonstration data sets of a variety of real world robots performing different manipulation tasks. The data set is increased through crowd-sourcing. It also provides a framework for generating demonstrations for a personal robot controlled by a phone through teleoperation.

\section{Applications}
\label{applications}

\begin{figure}[!t]
\centering
  \includegraphics[width=0.6\textwidth]{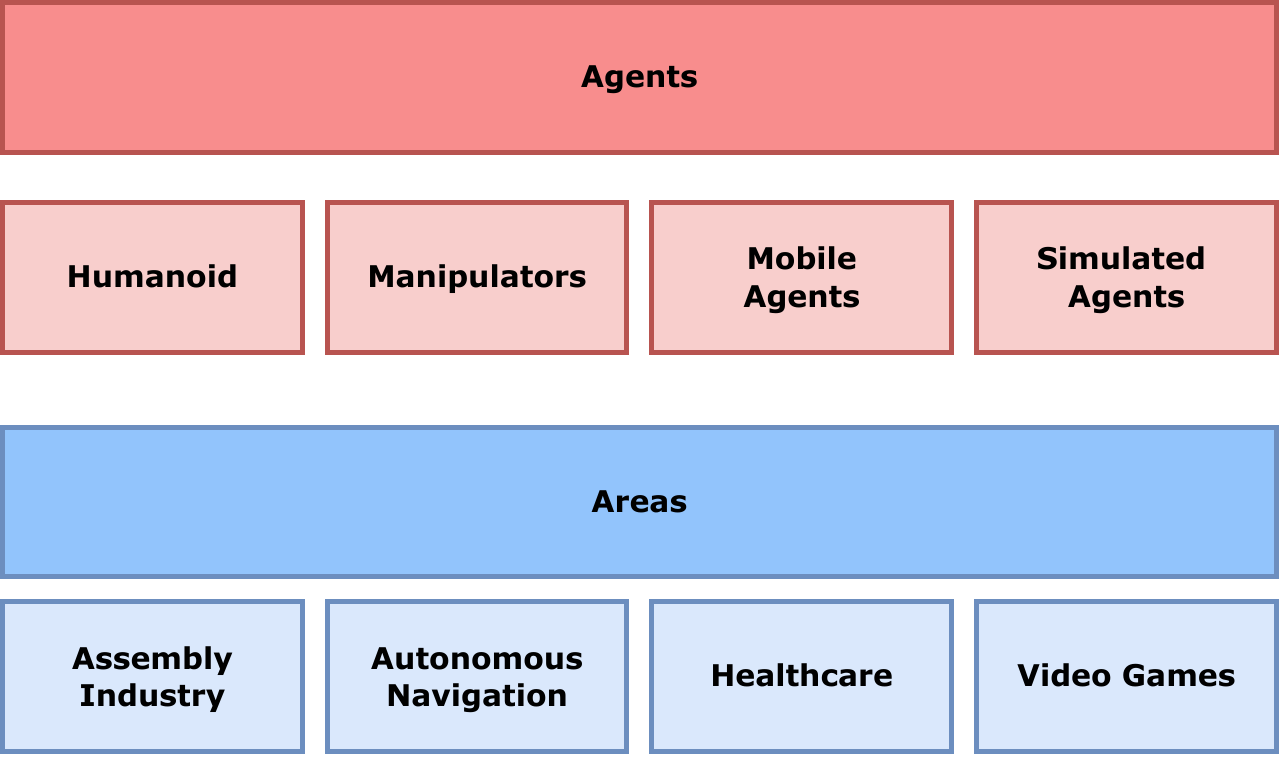}
    \caption{Agents and areas demonstration learning methods can be applied to.}
\label{applicationsfig}
\end{figure}

Demonstration learning approaches have been applied to a variety of areas as summarized in Fig. \ref{applicationsfig}. This section is dedicated to their exposition, complemented by a few relevant examples. The primary applications are robots and simulation environments. RL can learn complex skills but at the cost of interaction with the environment which can be dangerous or expensive in real world scenarios. Moreover, the sample inefficiency of RL requires millions of interactions even for simple tasks in the real world. For this reason, demonstration learning can be critical for applying policy learning algorithms to real world settings.

\subsection{Assistive Robots}
Demonstration learning can be applied to assistive robots which aim to help humans in their daily tasks. This is particularly difficult because the robot must adapt to all possible scenarios that the untrained human may require. 
In \cite{ikemoto}, the authors adapt the training process to account for the changing behavior of the human.
Robots may be used to interact with humans \cite{intervention} or help with social or mental problems \cite{bemelmans}. Assistive robots likely have to imitate how a human would assist, which is something that demonstration learning can provide over reinforcement learning.

\subsection{Autonomous Navigation}
The idea of a self-driving car has been one of the main focuses of machine learning in recent years. The goal is to use the information captured from a wide range of sensors to control the vehicle safely. Due to the dimensionality of the problem, reinforcement learning represents a viable solution by learning from interactions. However, the unsafe characteristics of reinforcement learning represent an obstacle, as the consequences of unsafe driving are catastrophic. Learning from demonstrations avoids this problem since the agent does not have to interact with the environment to learn.

Early research in \cite{sammut} proposes a method for learning to fly an aircraft from demonstrations recorded via teleoperation using IRL. Demonstration learning has also been applied successfully to autonomous aerial navigation such as drones for their potential for the delivery of goods, and helicopters. In \cite{helicopter} and in \cite{NASA}, data of a robot helicopter flight, using a joystick, was recorded and used to train an autonomous helicopter agent through reinforcement learning and apprenticeship learning, respectively.  
With advances in sensors for capturing data, demonstration learning has been increasingly used to learn control policies. Demonstration learning has also been employed for the locomotion of bipedal and quadrupedal robots. The demonstrations for training such robots are obtained either by teleoperation or observation. 
Some works have applied demonstration learning to driving tasks \cite{abbeel2004,smile,ziebart}.
In \cite{modataset}, a demonstration data set for navigation tasks is collected by a camera mounted on top of a robot.
In \cite{veloso}, the authors test their active learning based on demonstrations method on both simulation and real robot navigation. In \cite{saunders}, a mobile robot is used to capture information about its surroundings and choose the correct algorithm in its database to use in each situation. Full autonomous control of a vehicle in the real world is such a complex problem that it likely can not be solved using a single field of machine learning. Nonetheless, demonstration learning offers tremendous potential to learn skills from data without having to interact with the dangerous environment of driving in the real world. 

RobotCar \cite{maddern} and BDD100K \cite{bdd100k} are large data sets containing real-life driving demonstrations in the form of videos.
Demonstrations have been used towards autonomous driving in \cite{bojarski,pan,codevilla,bansal}. The safety learning from demonstrations can be used to employ RL in autonomous driving where the agent can not violate safety constraints learned from demonstrations \cite{rhinehart,constraint,safecont}.

\subsection{Manipulators}

Demonstration learning has been applied to manipulators in manufacturing applications since the 1980s. Training the manipulator using demonstrations, usually through kinesthetic teaching, is easier than hard-coding the behavior. A common application is assisting and healthcare robotics.
These are robots designed to help humans with tasks. Machine learning methods are used for such tasks because the robot must perform in any environment. Training a policy to generalize to new environments is easier than hard coding all the scenarios. Besides, assistive robots are expected to operate near humans. This increases the requirement for safety, which is increased with convergence and stability guarantees of the learned policy compared to hard-coded behaviors. Furthermore, effective collaboration requires the desired robotic movements to be similar to those of humans. Therefore demonstration learning is a good fit for generating such behaviors in the policies.

Several works have used demonstrations for learning robotic grasping \cite{pinto2016,levine2018}.
Some examples of demonstration learning methods applied to such robots include handing over objects to humans in \cite{handover}. The authors codify the coordination, represented in demonstration data, that allows humans to transfer the control of a specific object to other entities. 
A model-based algorithm \cite{ebert} learns a variety of manipulation skills.

The policies for learning manipulation tasks can replace humans in healthcare. However, this poses major concerns in safety where any exploration required by RL is impractical \cite{gottesman}.
In \cite{surgery}, the authors train multiple robotic arms to perform surgical tasks, where the trajectories were learned from demonstrations.
Later, \cite{tseng} applies a model-based method for the treatment of lung cancer.
The MIMIC-III dataset \cite{mimicdataset} provides 60 thousand ICU records and has been used for drug recommendations 
\cite{raghu} and to treat sepsis \cite{drug} by leveraging treatment processes demonstrated in the data set.
The automation of healthcare through demonstration learning can optimize the procedures and improve results.

\subsection{Humanoid Robots}
Humanoid robots are perhaps the most obvious application of demonstration learning. These are robots with a structure similar to humans whose goal is to perform a task instead of humans. Since the required tasks are already performed by humans, the concept of learning from demonstrations is suitable. Here the demonstrations can be captured through a sensor suit and converted into the values of the robot's joints. The tasks range from only using part of the robot \cite{vogt} to the entire humanoid body \cite{calinon,ude}.

In \cite{speert}, a humanoid robot was trained from human demonstrations captured from sensors placed directly on the human's body to perform reaching and drawing movements with one arm, and tennis swings. In \cite{ogino} a humanoid robot learns to imitate the arm gestures of a human demonstrator and is tested on a turn-taking gesture game. A human teleoperated a humanoid robot in \cite{NASA}. Using Virtual Reality technology to convert the operator's arm and hand motions into those of the robot, demonstration data was obtained to learn a manipulation policy. In \cite{tcn}, an embedding space was learned from demonstrations. The human demonstrations are projected into the embedding space, from which a humanoid robot is able to replicate the human movements.
In human-robot interaction, other than learning task control, the robot should learn where to focus its attention.

\subsection{Video Games}
Machine learning methods, mainly reinforcement learning, have been applied to video games successfully. Creating an agent for learning to play and beat the game or generating a controller for non-playable characters are the main applications of such approaches. However, video games often suffer from sparse rewards, and some may have relatively high dimensional spaces. This hinders the convergence of the policy during exploration. Furthermore, the behavior of non-playable characters often needs to appear human-like \cite{hingston}. Therefore, demonstration learning serves as a bridge to reach these goals that reinforcement learning struggles to achieve.

Demonstration learning has been applied to racing games such as Mario Bros in \cite{ortega} and \cite{mario}. The Atari Learning Environment \cite{ale} provides an environment for bench-marking algorithms on Atari games. Approaches such as \cite{deepq} have used this environment for evaluating their method and comparing it to other state-of-the-art methods for Atari games.
These examples are relatively simple problems, with small state and action spaces. However, they show a starting potential that can be extended to more complex games in the future.

The variety of genres and the continuous growth of the dimension of video game domains constitute difficult problems getting ever closer to the difficulty of the real-world. Methods often rely on simulation to validate their correctness, due to safety concerns of deploying the method in the real world. However, creating a simulator is difficult and often returns inconclusive results because the real world is still more difficult. Video games offer a diverse range of problems that demonstration learning methods can try to solve. Solving increasingly more difficult problems in video games will allow for a validation closer to the real world.

\section{Advantages and Disadvantages of Demonstration Learning}
\label{advdis}

\begin{table}[!t]
\caption{Summary of the advantages and disadvantages of demonstration learning methods.}
\label{advdistab}
\resizebox{\textwidth}{!}{

\centering

\begin{tabular}{l|l}
\multicolumn{1}{c|}{\textbf{Advantages}}                                               & \multicolumn{1}{c}{\textbf{Disadvantages}}                                                   \\ \hline
Diverse and flexible: can be combined with other paradigms.                                   & Creation of  the data set: difficult to create large numbers of high quality demonstrations. \\
Data efficiency: faster convergence from fewer data over reinforcement learning.       & Distributional shift: generalize from the data set or remain inside its distribution.        \\
Performance guarantees: the method will at least be as good as the demonstrations.     & Quality of the data set: performance relies heavily on the demonstrated behavior.            \\
Reduction of programming load: no need to program the model's decision for every case. &                                                                                              \\
Safety: the agent learns without interacting with the environment.                     &                                                                                              \\
Simplification: demonstrate the task and estimate a policy with a SOA method.          &                                                                                             
\end{tabular}
}

\end{table}

Different demonstration learning algorithms bring different advantages and disadvantages. These are summarized in Table \ref{advdistab}. This section focuses on the general advantages and disadvantages of demonstration learning methods.

\subsection{Advantages}
The clearest advantage of demonstration learning over other paradigms is the idea of not requiring an expert programmer. Even though the field has not reached this point, the goal is to train an agent to perform a task solely by providing demonstrations. The paradigm enhances adaptability, by allowing an agent to learn the behavior from demonstrations rather than being programmed with static instructions. It is much easier to demonstrate how to perform manipulation than to describe all the steps a robot has to do to perform it. The field allows agents to learn from external and different sources such as humans or other agents with different types of hardware. 

Additionally, the paradigm is very data efficient, especially compared to reinforcement learning which is severely data-inefficient. Several approaches use a small number of demonstrations. Reinforcement-learning methods learn through trial-and-error interactions to learn optimal policies. These errors are, for the most part, failed attempts. Demonstration learning allows the agent to learn the task by providing the correct choices for the input states. This property allows the paradigm to solve high-dimensionality problems and effectively addresses the curse of dimensionality machine learning problems suffer from. Furthermore, in reinforcement learning, the agent's policy can converge to a local minimum whose respective performance may be lower than required. In demonstration learning, the agent learns the behavior present in the demonstrations, meaning there are policy performance guarantees with this paradigm. Demonstration learning can then be paired with other machine learning fields, such as reinforcement learning that can improve the learned method.

Lastly, one of the main restrictions of reinforcement learning is that it is difficult to apply it to real-world problems. The agent learns from trial and error interactions, making it a perfect fit for simulation environments. However, in the real world, errors may have serious consequences. A collision may severely damage a robot, hindering or preventing learning altogether. 
Demonstration learning is extremely valuable in applications where interaction is impractical, expensive, or dangerous.
Demonstration learning provides the interaction data in the demonstration data set and the agent does not interact with the environment while learning. Therefore, the agent learns in a much safer way, allowing it to be applied to real-world problems. Even if interacting with the environment is safe, using demonstrations can speed up convergence and improve generalization in complex domains.
After learning a policy from demonstrations, the policy can still be improved with online interactions, with the benefit that because the policy has task knowledge, it is safer to interact with the environment than a random policy in RL.

\subsection{Disadvantages}
While learning from demonstrations is appealing and solves the issues of RL, it makes it challenging to learn from limited data. Technically, any off-policy RL algorithm can be applied to learn from the demonstration data set. However, these algorithms were designed for online RL where the agent can interact with the environment and correct its mistakes. Hence, these methods often fail when using limited data. 

The disadvantages of demonstration learning are related to the data set and were mentioned in section \ref{datalimit}. The quality of the policy is directly related to the quality of the demonstrations in the data set. Sub-optimal or incorrect demonstrations may hinder or prevent the policy from converging into adequate behavior. Existing solutions identify and filter these sub-optimal demonstrations. 

Additionally, demonstrations are expensive to collect and the data set may be insufficient, meaning that not all possible state actions are demonstrated. Creating the environment with the sensors, demonstrating, collecting the data, and creating adequate embodiment and record mappings for the data set, are all non-trivial steps.
Then, the distributional shift discussed in previous sections can severely hinder the performance of the model. The demonstration data set has a distribution that is a subset of the real MDP. Unless the data set is optimal the agent can not generalize to cover the entire problem. Hence, when queried in out-of-distribution states, it will likely fail. Restricting the agent to in-distribution states to avoid being queried for actions that the agent was incapable of generalizing to while learning from the limited data, is difficult and most methods directly address this issue explicitly.

\section{Future Directions}
\label{future}

\subsection{Bench-marking}
The goal of demonstration learning is to teach a real-world agent a task from demonstrations. However, most methods are evaluated in simulation environments.
Since demonstration learning is a recent area, there are not many standardized bench-marking environments and data sets, especially for real-world environments. Any research area can benefit from such standardization because it eases the comparison of different methods for the same problem, and demonstration learning is no exception. 
A real-world demonstration learning benchmark is an open problem in the field that is limiting its progression and applicability to real tasks.

Current demonstration learning research methods are often evaluated using a limited range of reinforcement learning environments and often create their own demonstration data set.
The method should be able to generalize to other environments and perform the task with high repeatably. The method should also be computationally efficient, and some cases require the behavior to appear human. 

Although some environments provide a respective demonstration data set, methods that use their own create uncertainty about the quality of the method. The number of demonstrations in the data set affects the performance of the method. Demonstrations are not easy to collect in real-world scenarios, hence the demonstration data sets do not tend to be large. However, some methods do use a large number of demonstrations. A larger number of demonstrations corresponds to a larger coverage of the state-action space and is a clear advantage over methods that use a smaller data set. The quality of the demonstrations is difficult to measure and compare between data sets. However, it is safe to assume two different data sets will have different quality levels. The performance of the method is proportional to the quality of the demonstrations.

Existing stochastic environments are very limited. Additionally, non-stationary environments, which are very common in the real world, are also very limited with the only exception being the Real-World RL suite. Lastly, from our knowledge, multi-agent environments and data sets are not available.
Because of these reasons, future works should create more standardized environments, demonstration data sets, and metrics and encourage their usage.

\subsection{Context Problem}

The context is the set of characteristics that define the environment. Background objects, illumination, and camera positioning are examples of such characteristics.
Most demonstration learning methods focus on learning a policy in a single context. Changing any of these elements results in a different context. Then, because the context is not the same as the one where the policy was trained to perform, the policy will likely not be able to perform the task in this new context, rendering it and all the work that went into its estimation, useless. This problem is usually swept under the rug because the evaluation of the methods is done in simulation or controlled real-world environments, allowing for the context to remain unchanged from training and the policy to successfully perform the task. This problem severely impacts the scalability of the policy in real applications where the context changes can be drastic.

\subsection{Goal Specification}

Demonstration learning methods focus on learning one specific task at a time. Pick-and-place tasks, for example, involve picking up an object from one location and placing it somewhere else. Picking a different type of object and/or placing the object in a different location is usually treated in the literature as a different task from the original \cite{squirl}, even though it is still a pick-and-place task. Learning a new task every time the goal changes means that the work required to have the agent perform all the required tasks scales linearly with the number of goals. Since these are technically the same task with different goals, there's redundancy and shared information between the policies. Future directions should train policies to scale and adapt to different goals specified in the input in addition to the state, instead of training a policy for each goal.

\subsection{General Demonstration Learning Framework}
Demonstration learning's end goal is to allow the training of an agent without expert programming knowledge.
An intelligent agent is trained solely by demonstrating a task.
However, current approaches still require the design of algorithms for feature extraction, reward specification reliant on the type of agent and task, and policy derivation algorithm. A learning framework that could be applied to any task would allow a teacher to train an agent solely by demonstrating the task.

\subsection{General Feature Extraction}
All demonstration learning approaches depend on the data set. The most practical form of representing a high-dimensional environment is through images. Learning from images requires the extraction of quality features that currently depend on the task. Creating a general feature extractor framework would eliminate the need for engineered feature extraction frameworks for each task. This is one of the open problems preventing the creation of a general demonstration learning algorithm.

\subsection{Generalization}

The policies should be able to select the correct action for any possible state. In demonstration learning, the policies are estimated from the demonstration data set. Hence, the policy is limited to the state-action distribution of the data set. Generalization is the ability to predict correct decisions for unseen scenarios based on previous scenarios. Some approaches increase the data set through techniques such as data augmentation \cite{svea,rad}. Others fine-tune the policy on online data \cite{coarse}. Alternatively, some approaches restrict the policy to in-distribution states such that it is never queried for actions it has not learned \cite{constraint}.

\subsection{Hyper-parameter Selection}
Hyper-parameters are common to most demonstration learning approaches. These parameters need to be specified beforehand. Such examples include choosing the number of hidden layers in a neural network and the number of neurons in each layer. 
Their ideal values are hard to estimate and are usually updated by trial and error. 
In \cite{showingwork}, the authors note that the results shown by most works are not sufficient to evaluate the methods because such results are based on optimal hyper-parameters and applied to a limited range of tasks. Instead, the methods should be trained multiple times varying the set of hyper-parameters and the resulting set of policies should be evaluated using their proposed metric named 'Expected Validation Performance'. 

Generally, it is difficult to perform policy rollouts in real-world settings, due to costs or safety concerns. This is especially true for evaluating policies that are still being trained. Performing evaluation before the training of the policy to determine the ideal value of hyper-parameters, would allow for changing the values of hyper-parameters and safe time and resources, instead of training a complete policy.
Hence, this is another reason restricting demonstration learning to simulators. Off-policy methods aim at evaluating a policy from previous data, without requiring interaction with the environment.
In \cite{voloshin}, a study is performed evaluating thirty-three different OPE methods by measuring the distance between the estimated policy value and the true policy value. The study concluded that evaluation with the state-action function outperforms the remaining methods. To help accelerate the development of OPE methods, \cite{fu} propose the Deep Off-Policy Evaluation (DOPE) benchmark. 
The available options are using inaccurate OPE methods or training the model for a fixed number of steps and adjusting the values based on the early results.
The challenge of the selection of hyper-parameter values is common to other machine-learning paradigms. However, demonstration learning needs to overcome this problem if it truly wishes to reduce the requirement of expert programming. We believe that the creation of general, accurate, and efficient OPE methods is fundamental for the growth of the field and expansion to more real-world applications.

\subsection{Long-Horizon Tasks}

Offline reinforcement learning relies on a reward function. The more frequent the rewards, the easier it is to estimate a policy. However, it isn't always trivial to create a reward function that provides a reward for every step. The problem is exacerbated when the task is complex. Hence, demonstration learning has been restricted to simple tasks. These are tasks that finish within a relatively small number of steps and require few interactions. Long-horizon tasks are more applicable to real-world scenarios. These can be a composition of small interactions or a single yet long and/or complex interaction. Because they require a larger number of steps to finish they are usually associated with a larger set of states for the agent to visit and learn. Even though learning long-horizon tasks remains an unsolved problem, some methods have tackled such tasks with some success.

The method in \cite{gti} detects that the demonstrations in the data set intersect at certain states with each other. With these intersections, it can generalize new trajectories.
\cite{ltsd, swirl} show that complex tasks can be divided into sub-tasks. The sub-tasks are learned from the demonstration data set. The agent receives a reward after reaching each sub-task. The authors of \cite{accelarating} define a skill as a useful sequence of actions and propose to learn an embedding space of skills from unstructured demonstrations. They then train a reinforcement learning policy on the embedding space using a hierarchical model where the high-level component generates embeddings of skills which the decoder then converts into sequences of actions. \cite{iris, relay} propose hierarchical frameworks composed of high-level and low-level policies. The high-level policy sets sub-goal states for the low-level to try and reach. In \cite{ris} the high-level policy is encouraged to predict intermediate states between the current state and the goal state.

RUDDER \cite{rudder} redistributes the reward by identifying key steps in the demonstrations and increasing the reward of the respective transition. However, RUDDER uses LSTMs to predict the associated reward and models require many demonstrations to generalize well. Because of this Align-RUDDER replaces the LSTM model with a profile model adapted from bio-informatics.
This model can be estimated with very few sequences. Then, aligning a sequence with the model generates an alignment metric for how compatible the sequence is with the model.
The reward of a transition is the difference between the alignment values of the sequence with and without the transition.

\subsection{Multi-Agent Demonstration Learning}
Most demonstration learning research is focused on single-agent learning. However, in real-world tasks, agents would ideally interact with each other and learn to cooperate. Even though some work has been done in multi-agent cooperation or competition, as discussed in the respective section, the applications of the methods are limited. Hence, multi-agent demonstration learning remains an open problem.

\subsection{Learning from Sub-Optimal Data}

Demonstration learning is fully dependent on the quality of the demonstration data set. However, in most settings, the data is sub-optimal due to a variety of reasons. 
The data captured from sensors is prone to inherit noise and sensor errors. These increase the distributional shift between the estimated models and the demonstrated behavior. Some works have tackled inadequate data through identification and removal \cite{toiletpaper} or correction \cite{repair}.
Gathering demonstrations is generally expensive, leading to a small number of demonstrations in the data set. To counter this, some methods choose to use demonstrations from other tasks allowing the policy to not only generalize better to a single task than if only the task's data set was used, but also causing the policy to be quickly adapted to other tasks. In \cite{corro}, data sets are collected from multiple tasks sharing the same state and action spaces, but varying reward functions and dynamics. An encoder is trained to encode trajectories into an embedding space. The policy is then conditioned on a task embedding in addition to the state. However, using data sets from other tasks can increase the distributional shift between the data set and the policy. Because of this, \cite{conservativedatasharing} proposes a method to learn policies from data sets of multiple tasks without causing the distributional shift, by relabeling the transitions from the data sets of the other tasks. 

Additionally, the estimated distribution of the policy might differ from the demonstrator's distribution. The larger the distributional shift, the worse the policy will generalize. In \cite{provablyeff}, the authors tackle the mismatch between the distribution of the offline data set and the distribution of the learned policy by deriving a trackable bound of this mismatch. This estimation is then used as an extra regulating parameter in the optimization step.

Offline RL requires reward specification which is expensive to provide in the real world. Other than IRL, a promising direction is to use unsupervised techniques to leverage unlabeled data.
In \cite{leverage}, a policy is learned from large sets of unlabeled data when combined with a smaller set of labeled data. In \cite{yarats}, a downstream reward labeling method is applied to unlabeled data sets.

\subsection{Safety Concerns}

Demonstration learning algorithms learn a policy to maximize a goal. In offline RL, the goal is to maximize the expected accumulated rewards, while in behavior cloning the goal is to maximize the correspondence of the action choices to the 
ones of the demonstrator. However, these goals don't take into account safety concerns. The demonstrator likely does not demonstrate all the possible safety-critical scenarios and the reward function likely does not take into account safety. Safety is a requirement for real world applications, especially in human-robot interaction settings \cite{desantis,ikemoto}.

In RL, existing algorithms tackle safety by specifying constraints that the agent can't violate during execution \cite{cbf,constraint,safecont,recovery}. However, these constraints are specific to a certain task and agent, and specifying all the constraints for each situation is impractical. Alternatively, \cite{intervention,androids} requires access to the set of safe and unsafe states to train the agent in a safe manner.
\cite{copo} starts by finding a reward-optimal policy first, this policy is then projected to the feasible set of policies that satisfy the cost constraints.

The DAgger framework trains a student policy using an expert policy that the student can query \cite{smile}. The student policy is trained on data generated by itself and the expert. Before each interaction with the environment, a decision rule decides which of the two policies to use. Selecting the student's action is unsafe early, hence SafeDAgger \cite{safedagger} adds an extra criterion where the student is only allowed to act if the difference between its action and the expert's action is below a threshold. Later, EnsembleDAgger \cite{ensembledagger} added an extra criterion where the prediction of an ensemble of networks must also be below a threshold for the agent to act. The downside of these algorithms is the requirement for an expert policy.

\section{Conclusion}
\label{conclusion}
We presented a comprehensive survey on demonstration learning. Demonstration learning reduces the overhead of programming by teaching the agent a task through demonstrations. The paradigm can be divided into two phases. The first phase consists of collecting and building the demonstration data set. There are many design choices to account for: choosing the demonstrator, recording the demonstrations, representing the data in the data set, and the common data set limitations and how to counter them. The second phase consists of extracting the behavior represented in the data set and training an agent.

The main challenge in demonstration learning is the generalization to unseen scenarios. Demonstrations only cover a subset of the distribution. Direct imitation through behavior cloning learns this distribution. If the agent visits out-of-distribution cases, not only will it not know what to do, but the consequences in the real world can be catastrophic.
Because of this, offline RL methods try to specifically reduce the distributional shift. 
Different ways of achieving this and mitigating the challenges were discussed. We also discussed model learning methods that try to tackle this problem by allowing the agent to generate new transitions without interacting with the environment. Then, methods for refining the behavior were also explored. These range from trial and error interactions with the environment using reinforcement learning, transferring the skill from one task to another using transfer learning, actively querying the teacher, or using evolutionary algorithms for optimizing the behavior. 

Demonstration learning is primarily applied to different types of robots and video games which can be used in different domains such as healthcare and industry. The advantages of demonstration learning include reducing the requirement for expert programming, it being more data-efficient than reinforcement learning and the agent remaining safe during the learning process. However, it requires a good framework for creating the demonstration data set and estimating the policy.

Demonstration learning is a recent area with a promising future. The main open areas for research were identified. 
Specifically, benchmarks and data sets should be the primary concern for the success of the field. As noted in \cite{levinesurvey} much of the success of deep learning is due to large data sets. Even recent applications still utilize relatively old methods paired with improved models and large data sets. As explained in this survey, many of the problems of current demonstration learning methods are due to inadequate data sets. Creating more, better and larger data sets or pipelines for generating such data sets is the fundamental direction for the success of the field.

\section*{Acknowledgements}
This work was supported by NOVA LINCS (UIDB/04516/2020) with the financial support of 'FCT - Fundação para a Ciência e Tecnologia' and also through the research grant ‘2022.14197.BD’.

\end{document}